# An Outline of Prognostics and Health Management Large Model: Concepts, Paradigms, and Challenges

1st Laifa Tao[1,2,3,4], 2nd Shangyu Li[1,2,3,4], 3rd Haifei Liu[2,3,4], 4th Qixuan Huang[2,3,4], 5th Liang Ma[2,3,4], 6th Guoao Ning[2,3,4], 7th Yiling Chen[2,3,4], 8th Yunlong Wu[2,3,4], 9th Bin Li[2,3,4], 10th Weiwei Zhang[2,3,4], 11th Zhengduo Zhao[2,3,4], 12th Wenchao Zhan[2,3,4], 13th Wenyan Cao[2,3,4], 14th Chao Wang[2,3,4], 15th Hongmei Liu[2,3,4], 16th Jian Ma[1,2,3,4], 17th Mingliang Suo[2,3,4], 18th Yujie Cheng[2,3,4], 19th Yu Ding[2,3,4], 20th Dengwei Song[2,3,4], 21st Chen Lu[1,2,3,4,*]

[1]Hangzhou International Innovation Institute, Beihang University, China

[2]Institute of Reliability Engineering, Beihang University, Beijing, China

[3]Science & Technology on Reliability & Environmental Engineering Laboratory, Beijing, China

[4]School of Reliability and Systems Engineering, Beihang University, Beijing, China

*Corresponding author, luchen@buaa.edu.cn

1st taolaifa@buaa.edu.cn, 2nd lishangyu@buaa.edu.cn, 3rd phoebeliu@buaa.edu.cn, 4th qxhuang@buaa.edu.cn, 5th buaaml@buaa.edu.cn, 6th sy2214210@buaa.edu.cn, 7th 18373684@buaa.edu.cn, 8th wuyunlong@buaa.edu.cn, 9th libin1106@buaa.edu.cn, 10th 19375093@buaa.edu.cn, 11th zhengduozhao@buaa.edu.cn, 12th 20376049@buaa.edu.cn, 13th 20376030@buaa.edu.cn, 14th wangchaowork@buaa.edu.cn, 15th liuhongmei@buaa.edu.cn, 16th 09977@buaa.edu.cn, 17th suozi@buaa.edu.cn, 18th chengyujie@buaa.edu.cn, 19th dingyu@buaa.edu.cn, 20th songdengwei@buaa.edu.cn, 21st luchen@buaa.edu.cn

## Abstract

Prognosis and Health Management (PHM), critical for ensuring task completion by complex systems and preventing unexpected failures, is widely adopted in aerospace, manufacturing, maritime, rail, energy, etc. However, PHM's development is constrained by bottlenecks like generalization, interpretation and verification abilities. Presently, generative artificial intelligence (AI), represented by Large Model, heralds a technological revolution with the potential to fundamentally reshape traditional technological fields and human production methods. Its capabilities, including strong generalization, reasoning, and generative attributes, present opportunities to address PHM's bottlenecks. To this end, based on a systematic analysis of the current challenges and bottlenecks in PHM, as well as the research status and advantages of Large Model, we propose a novel concept and three progressive paradigms of Prognosis and Health Management Large Model (PHM-LM) through the integration of the Large Model with PHM. Subsequently, we provide feasible technical approaches for PHM-LM to bolster PHM's core capabilities within the framework of the three paradigms. Moreover, to address core issues confronting PHM, we discuss a series of technical challenges of PHM-LM throughout the entire process of construction and application. This comprehensive effort offers a holistic PHM-LM technical framework, and provides avenues for new PHM technologies, methodologies, tools, platforms and applications, which also potentially innovates design, research & development, verification and application mode of PHM. And furthermore, a new generation of PHM with AI will also capably be realized, i.e., from custom to generalized, from discriminative to generative, and from theoretical conditions to practical applications.

# 1. Introduction

Prognostics and Health Management (PHM) serves as an essential mechanism for a complex system to mitigate catastrophic incidents, ensure mission reliability, enhance operational efficacy, and curtail maintenance expenditures. By seamlessly integrating into the entire life cycle of a complex system, PHM has transpired as an essential fulcrum, wielding a pivotal influence reminiscent of a trump card. The application of PHM technology has increased the sortie rate of advanced fighters such as F-35 by 25%, reduced the maintenance manpower by 20%-40%, and reduced the use and support costs by 50% (Chen L et al., 2016). Furthermore, helicopters like the UH-60L *Black Hawk* experienced an elevation in operational readiness by 27%, a decrement in unplanned maintenance by 52%, and an overall decline in repair tasks by 17% (Baozhen Z & Ping W, n.d.). Large manufacturing conglomerates, such as Boeing, have also reaped the benefits of PHM by curtailing losses due to malfunctions by an impressive 15% (M D et al., 2016) Hence, the intrinsic value of PHM in fortifying mission success, elevating safety measures, paring down maintenance overheads, and amplifying efficacy within the industry cannot be understated.

Due to the influence of complex operating conditions and harsh environments, existing PHM methods face a series of impediments to achieving satisfactory results. Non-ideal characteristics of PHM data and knowledge(J. Peng et al., 2022; S. Wang et al., 2023; Zio, 2022), inadequate generalization capabilities of PHM models(Chu & Zhu, 2021; Tian et al., 2022), and the dependency of PHM capabilities on specific scenarios(Y. Sun, Lu, et al., 2023; Ye et al., 2020) have resulted in a plethora of pain points for current PHM technological approaches. Although potential interim solutions to a minority of these issues exist in contemporary research, these solutions often exhibit strong customization, weak generalization, and are accompanied by exorbitant costs. The improvements they bring are frequently marginal, failing to meet the capability demands of PHM technologies in application scenarios. Furthermore, the rapid emergence of various new systems and products, along with the application in new and intricate scenarios, exacerbates the conditions for PHM technological deployment. Yet, there are demands for multi-scale PHM capabilities that span from individual components to system clusters(W. Yang et al., 2021), as well as the rapid formation of PHM system capabilities. This juxtaposition leads to an inherent contradiction and conflict between traditional PHM methods and these new requirements. There is an urgent need to explore a new PHM methodological framework that holistically addresses the series of bottlenecks confronting PHM and fulfills the rapid capability formation demands of PHM in the current era.

Artificial General Intelligence (AGI)(Alattas et al., 2021; Iman et al., 2021; Nti et al., 2022) denotes an artificial intelligence that possesses intelligence equivalent to or surpassing that of humans, capable of manifesting all intelligent behaviors characteristic of average humans. It is quintessential to the realization of AI. In recent years, Large Model technologies, exemplified by Large Language Model (LLM) (Devlin et al., 2019; Openai et al., n.d.; Vaswani et al., 2017), have achieved remarkable strides, unveiling colossal potential towards the actualization of AGI. LLM employ generative modeling on vast textual corpora. When both dataset scale and model size reach a certain inflection point, they can exhibit astounding generative prowess(Ganguli et al., 2022). LLM, with ChatGPT(Brown et al., 2020; Lewis et al., 2020; Radford et al., n.d.) being a paragon, employ techniques like instruction alignment, reinforcement learning(Uc-Cetina et al., 2023), fine-tuning(P. Liu, Yuan, et al., 2023), and thought chain for training and adjustment(Wei, Wang, et al., 2022). This equips them with robust generalization, inference, decision-making, and generative capabilities. In applications like human-machine dialogue and Q&A, their performance rivals or even surpasses human-level competence. Preliminary applications of these models have been explored in specialized fields such as healthcare(J. K. Kim et al., 2023; Lokesh et al., 2022), finance(Pawan Kumar Rajpoot & Ankur Parikh, 2022; H. Yang et al., 2023), manufacturing(Lykov & Tsetserukou, 2023; L. Wu, Qiu, et al., 2023), and education(Jeon & Lee, 2023; Tsai et al., 2023).. Currently, the philosophy and technology behind Large Model are being researched and employed across myriad fields. They are upending conventional specialized field research and application paradigms, ushering in a novel technological revolution. This paradigm shift is profoundly transforming traditional technological field development patterns and the modus operandi of human production.

Given this context and leveraging the capabilities of generative Large Model – notably their inference(Zheng et al., 2023), robust generalization(Igl et al., 2019), and multi-modal information analysis(Lyu et al., 2023) along with the current efficacies of Large Model across various fields, there is a pressing need to address the bottlenecks and practical requirements pertaining to PHM technology. Aligning with global technological advancements, there's an imperative for in-depth integrative research between Large Model and PHM technologies. The overarching objective is to augment the core PHM capabilities and the proficiency of tasks throughout the entire life cycle of PHM systems. This would involve surmounting challenges associated with constructing, training, optimizing, and deploying PHM-LM (Prognosis and Health Management Large Model). By synergizing applications of Large Model, collaborating between ordinary models and Large Model, and emphasizing specialized field Large Model development, there's an opportunity to refine current PHM operational frameworks, enhance PHM algorithm competencies, and bolster downstream PHM tasks. This would, in turn, disrupt conventional PHM design, research and development, validation, and application paradigms. It paves the way for the inception of novel technologies, methodologies, platforms, and applications under the Large Model system framework for PHM. Ultimately, the ambition is to catalyze a paradigm shift in PHM technology: transitioning from bespoke to generic solutions, from discriminative to generative approaches, and from idealized settings to more pragmatic, real-world implementations.

In light of the practical demands associated with research on the convergence of Large Model and PHM, this paper integrates the technological advantages of Large Model, the systems engineering process throughout the entire life cycle of PHM, and the challenges faced by PHM. Building on a systematic review of the extant bottlenecks and challenges in PHM, coupled with a profound analysis of the merits of Large Model technology, we innovatively introduce a new concept of the PHM-LM. Additionally, we spotlight three prototypical innovative paradigms for future technical research and application advancements in this realm:

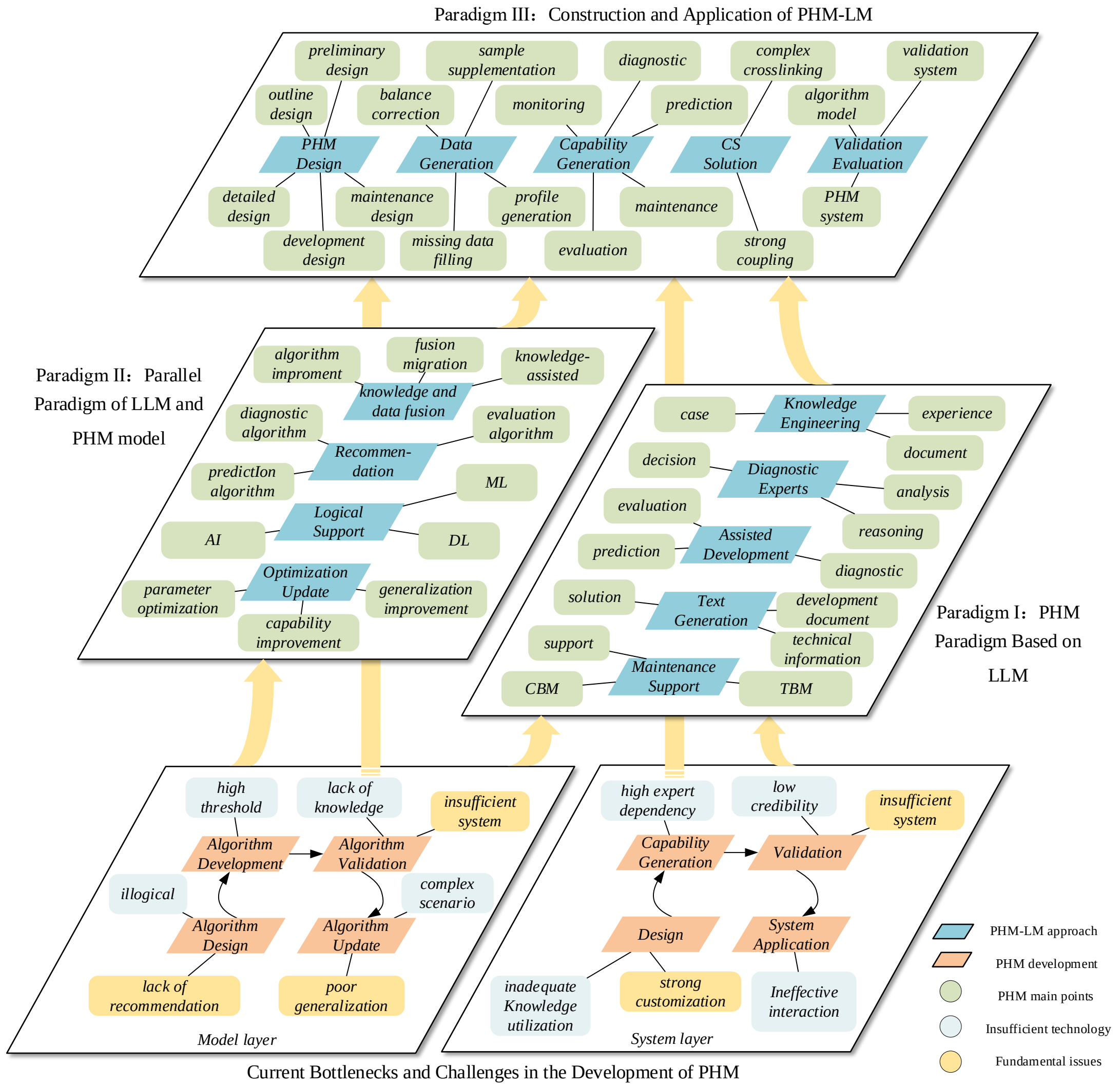

**Fig. 1-1 PHM-LM Progressive Paradigm Design**

Paradigm I: PHM Paradigm Based on LLM – Leveraging the linguistic reasoning and generalization capabilities inherent to foundational LLM, field-specific LLM are constructed via swift fine-tuning with PHM field knowledge.

Paradigm II: Parallel Paradigm of LLM and PHM Model – Grounded on the pretrained PHM models, a targeted field-specific learning process is ushered in, culminating in a collaborative development modality between the Large Model and the PHM models.

Paradigm III: Construction and Application Paradigm of PHM-LM – Given the unique characteristics of the PHM specialized field, there's a comprehensive and systematic design of the PHM-LM. This supports a plethora of downstream tasks, encompassing system PHM design, diagnosis, evaluation, forecasting, decision-making, recommendations, validation, and updates. Subsequently, under the new concepts and three typical innovative paradigms, this study delineates feasible technical routes by which Large Model can bolster key capabilities within PHM. Moreover, in addressing the bottlenecks and challenges inherent to PHM, the paper posits systematic solutions within the framework of Large Model. It further underscores a myriad of pivotal technologies and challenges that demand breakthroughs during the research and application phases of PHM-LM.

The aforementioned work, following the logical framework of demand-oriented, frontier-guided, and deep integration, systematically delineates and elaborates on the conceptual essence and technical paradigms involved in the integration research of Large Model and PHM. It also expounds on the paths to enhance core capabilities of PHM under the Large Model theoretical framework, technical means to address PHM challenges, and the technical hurdles of PHM-LM models. This provides clarity for readers regarding the current forefront of PHM technology and future development trends, and offers pivotal references for researchers engaging in the amalgamation of Large Model and PHM studies.

In the work under consideration, an integrative approach guided by demand orientation, frontier leadership, and profound integration has been employed. This study systematically elucidates the conceptual nuances and technical paradigms pertinent to the fusion of Large Model and PHM. Specifically, insights into the enhancement of core PHM capabilities under the ambit of Large Model theoretical structures, technical strategies addressing PHM impediments, and the technical quandaries associated with expansive PHM models are presented. This discourse offers a lucid orientation to contemporary frontiers in PHM technology and future trajectories, serving as an invaluable reference for scholars navigating the confluence of Large Model and PHM.

The subsequent structure of this paper is delineated as follows: Section II offers a comprehensive assessment of developmental bottlenecks and challenges encapsulating the PHM sphere. Section III encompasses an examination of the current landscape and intrinsic merits of Large Model research, augmented by an inclusive literature review.

In Section IV, a pioneering exposition of the PHM-LM concept is undertaken, highlighting three quintessential paradigms signifying the evolution of technological inquiry and applications in this field. Section V delves into the deployment modalities and application architectures of the PHM-LM. Section VI accentuates a constellation of pivotal technologies and challenges warranting breakthroughs within the purview of PHM-LM research and deployment. Conclusions are drawn in Section VII, encapsulating the salient points of the discourse.

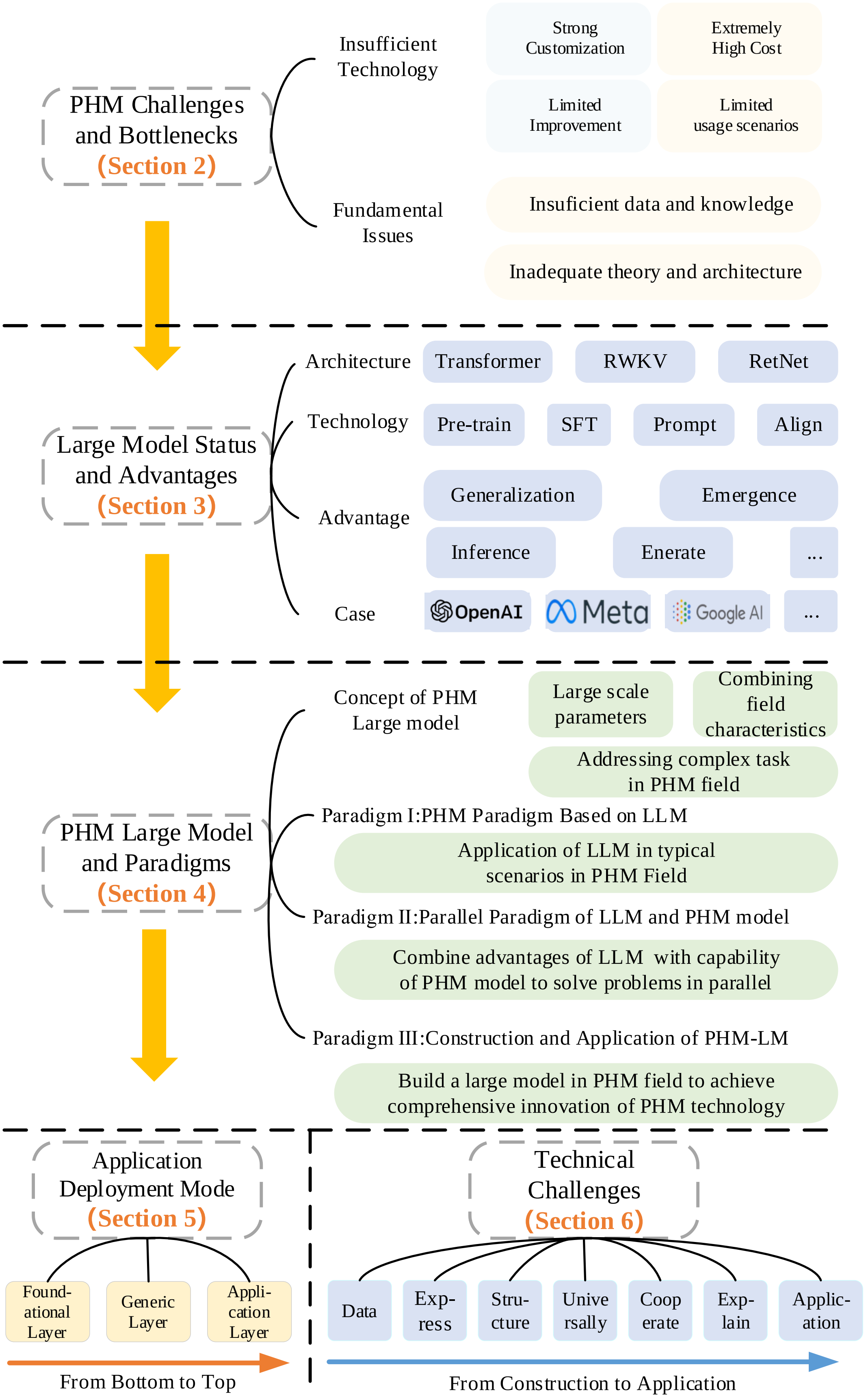

**Fig. 1-2 PHM-LM Outline Architecture**

# 2. Systematic Analysis of the Current Challenges and Bottlenecks in the PHM Field

In this paper, the PHM system is defined as one of the subsystems delivered with the system, possessing system-level functional tasks. Consequently, based on the real-world requirements of the PHM system and centered on the system engineering process (as shown in Fig. 2-1), this paper comprehensively analyzes the entire life cycle process of the PHM system(Ma et al., 2022; Tidriri et al., 2016). It systematically examines the series of significant issues faced by the PHM system and the PHM algorithm & model at different stages. Subsequently, the challenges confronting PHM are summarized and organized.

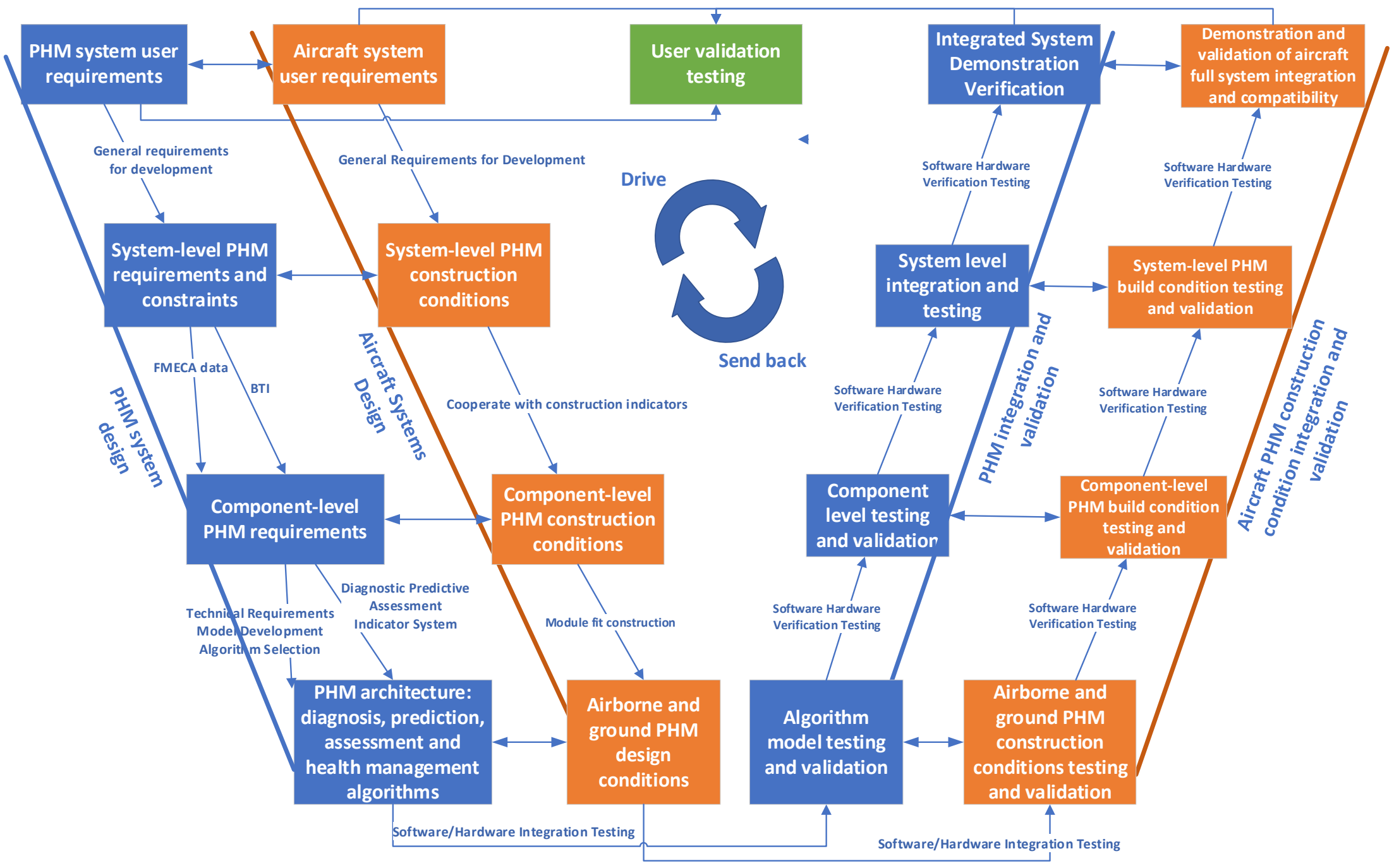

**Fig. 2-1 PHM System Dual-V Development Process**

## 2.1 Analysis and Description of PHM Issues Throughout Life Cycle

Surveying the development process structures of various advanced PHM systems, the entire life cycle of a PHM system can be broadly divided into five major phases. This section analyzes the essential tasks and work items to be completed in each phase and systematically examines the unresolved issues and capabilities that need enhancement at each stage.

**Phase One: Conceptual Design Phase**

During the conceptual design phase, preliminary PHM system design concepts are formed based on PHM design requirements. These concepts undergo validation and evaluation tasks, resulting in reports related to the PHM system development validation and evaluation analysis of the PHM system validation. This phase will produce overarching validation reports for the PHM system, including architecture validation, component validation, and software design validation. Key technological maturity, risk analysis and countermeasures, safety, and economic indicators of the PHM system are evaluated to complete the conceptual design. The design and validation work in the conceptual design phase requires a wealth of PHM field knowledge as support. The efficiency of knowledge utilization is a significant factor affecting the efficiency of this phase.

**Phase Two: Preliminary Design Phase**

In the preliminary design phase, based on the initial overall design, subsystem design, support solutions, and data/software system design, the core PHM functions such as testability, monitoring, detection, localization, isolation, evaluation, and prediction are designed. Subsequently, interface dependencies are allocated and docked based on the design content, completing the overall functionality, safety, economic, and risk assessments, laying the foundation for the detailed design of the PHM system.

**Phase Three: Detailed Design Phase**

During the detailed design phase, work is carried out on the technical solution of the PHM system, functional allocation prediction, etc., considering experimental research on the principles and functions of the PHM system. The detailed design is evaluated based on the experimental validation of principles and functions, and the prototype design of the PHM system is completed, including the prototype, simulation, and validation platforms.

**Phase Four: Development Phase**

After integrating the detailed design of the PHM system and the development standards and specifications, the construction of the PHM system starts from the underlying algorithm and gradually extends to the component level, sub-system level, and system level. Technical documents are written, providing related technical materials, help documents, and training materials for the PHM system.

**Phase Five: In-Service Phase**

During the in-service support phase, the PHM system serves alongside the system, realizing its PHM functions. As a large amount of usage data is generated, many iterative updates and new requirements will emerge, and the PHM system will need to undergo gradual iterative updates during the in-service support process.

## 2.2 Bottleneck Analysis of PHM

Based on the analysis in Section 2.1, to address the critical issues in the PHM field, break through the technological bottlenecks of PHM, verify and enhance the capabilities of PHM systems, and promote comprehensive development in the PHM field, this paper elucidates the challenges still faced by PHM technology from two perspectives: the PHM algorithm & model layer and the PHM system layer. This approach provides references at different levels for readers with varying research interests and needs, as illustrated in Fig. 2-2.

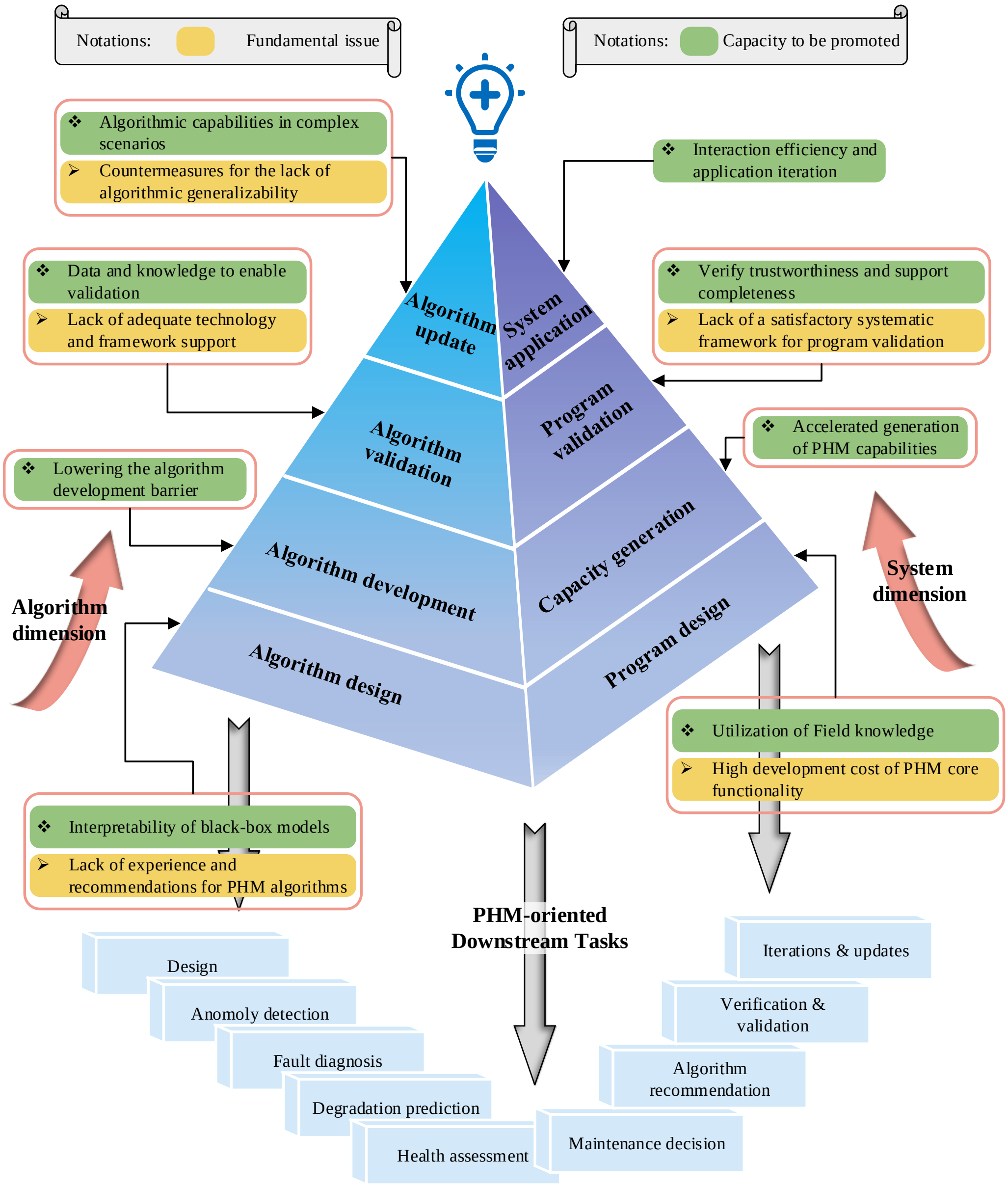

**Fig. 2-1 Bottlenecks and Challenges in the Development of the PHM Field**

### 2.2.1 PHM Algorithm & model Layer

**Challenge 1: How to lower the development threshold for algorithms & models?**

Advanced AI technologies and big data processing capabilities have brought significant improvements to the capabilities of the PHM field. However, due to factors such as object differences, data constraints, knowledge disparities, and variations in usage environments and conditions, the threshold for developing customized PHM models is gradually increasing. Coupled with the explosive growth of current PHM algorithms & models, tasks such as multimodal data processing design, algorithm & model selection, underlying code writing, and model hyperparameter configuration and iterative updates pose higher technical requirements for users. There lies a key challenge for the widespread application of the PHM field on methods to achieve optimal algorithm & model recommendations, automatic generation of underlying code, and adaptive optimization updates of model hyperparameters to lower the PHM model development threshold.

**Challenge 2: How to effectively carry out algorithm & model validation?**

In the PHM field, the capability validation of algorithms & models is crucial for ensuring PHM capabilities. However, the lack of data and knowledge during the design and development phase severely restricts the progress of validation tasks. Against this backdrop, how to innovate validation approaches and methods, establish auxiliary conditions for validation, and form a standardized, efficient, and trustworthy PHM algorithm & model validation system is a significant challenge that the PHM field urgently needs to address.

**Challenge 3: How to enhance the logicality and credibility of algorithms & models?**

The low credibility of the black-box algorithms & models widely used in the PHM field is an undeniable shortcoming in the field. With ample field-specific knowledge support, figuring out how to supplement black-box algorithms with explanatory methods in the PHM field, thereby enhancing the logicality and credibility of algorithms & models, is a key challenge for the development of high-trust PHM.

**Challenge 4: How to improve the generalization performance of algorithms & models?**

With the technological advancement in the PHM field, the requirements of PHM are continuously increasing. The challenge lies in identifying the common characteristics of different object PHM algorithms & models and the inherent common knowledge among numerous objects and systems. This ensures that the PHM algorithms & models maintain a high precision level across different objects, conditions, data formats, and constraint scenarios. Researching and developing PHM algorithms &

models with strong generalization and good generalization, or exploring technical methods to ensure and enhance the generalization capability of PHM algorithms & models, is a key challenge for the widespread adoption of PHM technology in the future.

### 2.2.2 PHM System Layer

**Challenge 5: How to efficiently utilize knowledge in the PHM field?**

With the development of the PHM field to date, the reliance on expert experience has become increasingly pronounced. The role of PHM knowledge spans various stages, including the refinement of PHM system design requirements, preliminary design (including algorithm & model design), detailed design (including algorithm & model design), algorithmic model validation, functional validation, and system integration validation. In an environment where multi-modal and multi-product data and knowledge are rapidly expanding, the challenge lies in enhancing the utilization efficiency of diverse knowledge and multi-modal data, replacing expert individual decision-making with question-answering reasoning based on generative model technology, increasing the connections between vast document knowledge, and consolidating fragmented knowledge with multi-modal data to form a comprehensive PHM knowledge base. This is a significant challenge for the future inheritance and development of the PHM field.

**Challenge 6: How to lower the design threshold for PHM systems?**

PHM system design often proceeds in parallel with the main system design and is subject to various constraints such as resources and deployment. Similar to PHM algorithm & model design, the design work becomes particularly complex due to constraints like monitoring objects, data transmission conditions, and computational resources. However, given the rapidly growing demand for PHM system development, relying solely on expert experience for PHM system design can no longer meet the efficiency and quality requirements of PHM system development. The challenge is to assist in PHM system design, construct generative PHM system design, lower the PHM system design threshold, and enhance the efficiency and quality of PHM system design.

**Challenge 7: How to establish a PHM validation system?**

In the PHM field, the validation and determination of solutions are crucial for ensuring that the PHM system achieves its intended functions and is developed with low consumption and high efficiency. However, the lack of data and knowledge during the design and development phase severely restricts the progress of PHM system validation tasks. Against this backdrop, the challenge lies in innovating validation methods, establishing auxiliary conditions for validation, and forming a standardized, efficient, and trustworthy PHM validation system.

**Challenge 8: How to enhance the basic performance and generalization of PHM systems?**

With the technological advancement in the PHM field, the basic performance requirements for PHM system are continuously increasing. The challenge is to identify the common issues of different object PHM systems, avoid the awkward situation of having to develop a PHM system from scratch for different objects, and design a PHM system with strong generalization capacity that is widely applicable to various systems.

**Challenge 9: How to enhance the interaction efficiency and performance of PHM systems?**

With the emergence of intelligent application terminals and diversified interaction forms, there’s vast potential for the development of application forms and human-machine interaction models in the PHM field software platform. Simultaneously, user feedback from human-machine interactions and the vast generation of real data provide an essential foundation for optimizing and updating the performance of PHM algorithms & models during their service phase. Breaking away from the current traditional human-machine interaction forms, innovating intelligent PHM applications, and exploring feasible solutions for the adaptive optimization and enhancement of PHM algorithm & model performance during their service phase based on human-machine interaction feedback and real data are significant challenges for technological innovation in the PHM field.

## 2.3 Summary

Starting from the PHM system engineering process, this section systematically analyzes the various issues that PHM still faces in each development phase and summarizes the bottlenecks and challenges encountered. These challenges and problems can be categorized into two scenarios:

(1) Scenario One: Issues such as the lack of generalization in PHM solutions and algorithms & models, low utilization rate of field knowledge, and insufficient optimization of the PHM system. Although there are some solutions or alternative approaches to these issues at present, the methods to address them are highly customized. They have weak generalization capabilities and often come with very high costs. The improvements they bring are often limited, necessitating further research and optimization.

(2) Scenario Two: Problems like the difficulty in conducting validations due to a lack of data and knowledge, and the poor interpretability resulting from the extensive use of black-box models, are currently insurmountable given the existing technological conditions. These issues are significant impediments to the development of the entire PHM field. Failing to address these problems will severely impact the future growth of the PHM field. It's imperative to innovate from a technological perspective and fundamentally revolutionize the theoretical framework.

Building on this foundation, Section 2.2 approaches from two different perspectives: the PHM algorithm & model level and the PHM system level. It identifies 9 major challenges facing the future development of the PHM field, focusing on knowledge utilization, logical supplementation, generalization capability, validation systems, development simplification, and interaction innovation. This is done to address more practical problems in PHM engineering and pave the way for technological innovation in the PHM field.

# 3. Analysis of The Current Status and Advantageous Features of Large Model Research

Large Model (LM) is a type of artificial intelligence model that imports massive amounts of unlabeled data into models with billions of parameters for large-scale pre-training, thereby training the model to adapt to various downstream tasks. Large Model can be divided into general field Large Model and specialized field Large Model. General field Large Model refers to large-scale pre-trained models handling multiple fields and tasks, while specialized field Large Model targets specific sectors or industries offering higher expertise and better performance for specific tasks. Currently, the most widely used Large Model is the LLM in the general field. This chapter, based on introducing the main development trajectory of Large Model, focuses on analyzing advantages and features as well as new technologies and ideas in the processes of construction and application. In particular, it elaborates on the advantages of the LLM in Large Model, analyzing their feasibility and rationality when applied to the field of PHM. All in all, this part sets the foundation for a seamless integration with EHM and addresses a series of issues in PHM.

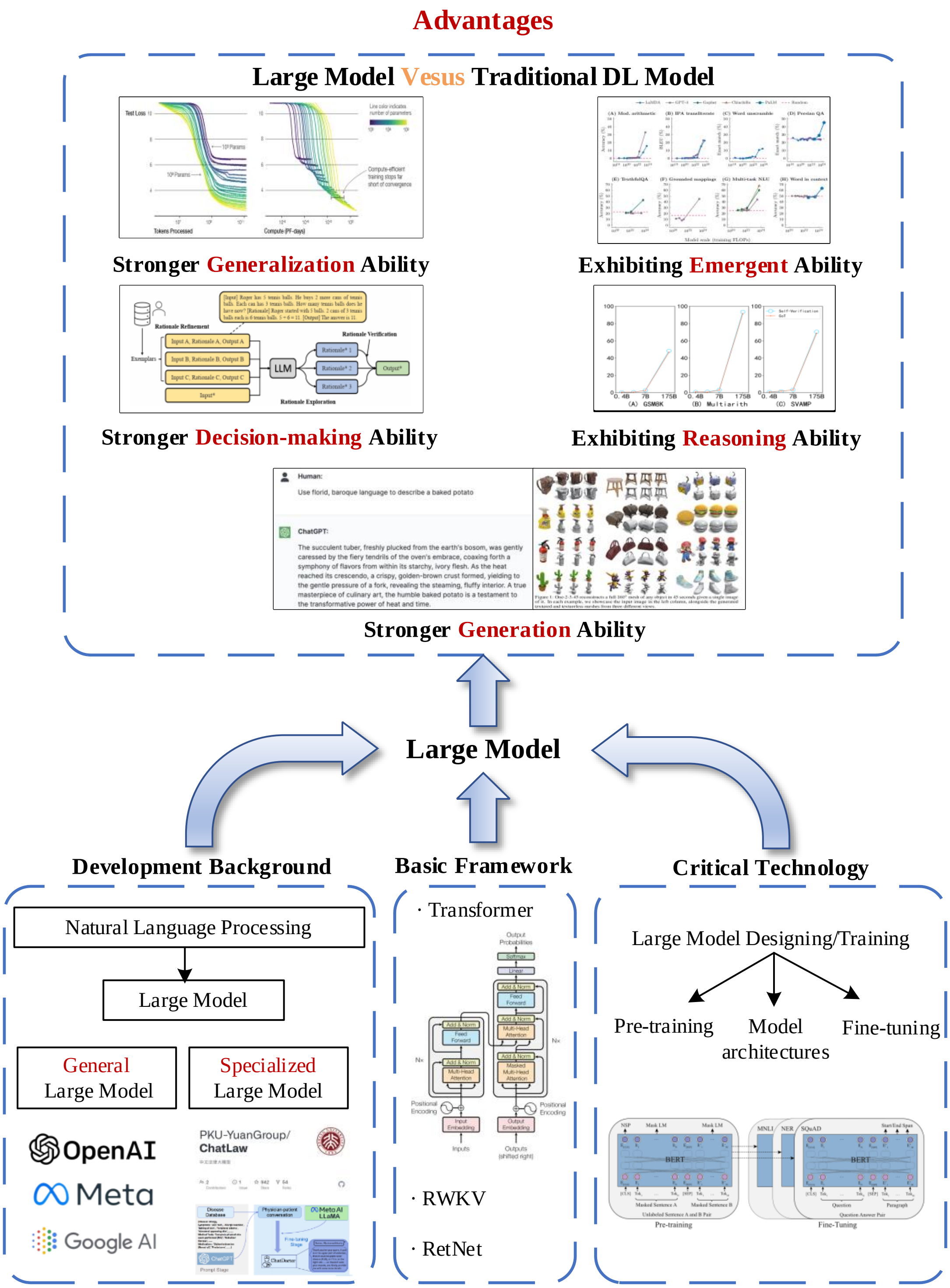

**Fig. 3-1 Analysis of the Current Status and Strengths of Large Model Research**

## 3.1 The Overview of Large Model

### 3.1.1 Large Model Principle

#### 3.1.1.1 Introduction to Large Model

With the rise of deep learning and the enhancement of computational capabilities, researchers have begun to focus on how to build models of larger scale to tackle more complex problems. Consequently, the concept of Large Model was introduced and has gradually become a focal point of development in today's technology landscape(Lourenco & Eugenio, 2019). In the era of IT, with data volumes growing explosively daily and the computational power of computers continually increasing, a key research direction in computational science today is how to handle unstructured and multi-modal data(Lipnicka & Nowakowski, 2023; Mayer-Schönberger & Cukier, 2013). These *data revolutions* also enable Large Model to have access to richer training resources and to process increasingly intricate data. In 2017, Google introduced the *Transformer* architecture based on the self-attention mechanism(Vaswani et al., 2017) scaling up to hundreds of millions of parameters. Building on the Transformer, in 2018, the introduction of the *BERT* model(Devlin et al., 2018) saw Large Model breaking the 300 million parameter barrier for the first time. With the emergence of the GPT mode(Brown et al., 2020), Large Model has garnered growing attention.

Large Model utilizes massive data to train an extensive number of parameters, thereby handling complex, abstract tasks that traditional ML and DL can't achieve. The objective of designing Large Model is to enhance their representational capabilities and performance, capturing data patterns and regularities more effectively when dealing with intricate tasks. Initially, Large Model was intended to solve issues in natural language processing (NLP). However, as it continued to evolve, its application has expanded far beyond just the NLP realm. Large Model can fit training data better, thereby enhancing its accuracy and generalization capabilities. Additionally, it can learn more intricate features, leading to better performance on complex tasks. As a result, Large Model has found applications in various fields, achieving significant breakthroughs in areas such as NLP(X. Wang et al., 2023), computer vision(Kirillov et al., 2023), voice recognition(Y. Zhang et al., 2023), and even in fields like law and medicine(Lu et al., 2021). In short, Large Model is playing an increasingly pivotal role in contemporary technology.

#### 3.1.1.2 Large Model Basic Architecture

Currently, the majority of Large Model adopt the Transformer architecture, which overcomes the shortcomings of RNNs, previously the most commonly used in the field of NLP. As a result, Transformers are widely utilized by numerous LLM and another Large Model. Apart from the extensively used Transformer architecture, new architectures, such as RWKV, have been proposed, offering fresh perspectives for the construction of Large Model.

(1) Transformer

The Transformer is the most popular architecture used in Large Model today, proposed by Google researchers in 2017. The Transformer comprises Encoder and Decoder modules. The encoder consists of a self-attention layer and a feed-forward neural network, employing residual connections and layer normalization within the layer. The encoder serves to transform input sequence data into a set of representation vectors. Meanwhile, the decoder also uses a combination of the self-attention mechanism and feed-forward neural networks, incorporating two attention layers and one feed-forward network, also employing residual connections and layer normalization. The decoder is responsible for outputting the predicted sequence. The Transformer effectively addresses the long-term dependency issue, enhances the model's parallelism, is broadly applied across various fields, and laid the foundation for models such as BERT and GPT(Devlin et al., 2019; Y. Liu et al., 2019; Openai et al., n.d.).

(2) RWKV

RWKV is an improved recurrent neural network (RNN) architecture. It overcomes the traditional drawbacks of RNNs that make them challenging to train on long sequences due to the vanishing gradient problem, as well as the Transformer architecture's downside of consuming a significant amount of memory during computation(B. Peng et al., 2023). Embodying the advantages of both the Transformer and RNN, RWKV can be efficiently trained and also support fast inference, making it a valuable architecture in the current landscape.

(3) RetNet

Jointly introduced by Microsoft Research and Tsinghua University(Y. Sun, Dong, et al., 2023), RetNet is a novel architecture composed of multiple identical blocks. Each RetNet block encompasses a multi-scale retention (MSR) module and a feed-forward neural network. Furthermore, RetNet introduces a multi-scale retention mechanism as a substitute for the self-attention mechanism, supporting both parallel and recurrent computation modes. RetNet demonstrates excellence in long-sequence modeling, and its implementation and deployment are relatively straightforward. It provides a new direction and breakthrough in the design of Large Model architectures.

3.1.1.3 Large Model Key Technology

(1) Pre-training Data Collection

To train Large Model, knowledge needs to be learned from vast amounts of data and stored in the model's parameters. In order to enhance the efficiency of model training, high-quality data must be used(Penedo et al., 2023). Given the significant presence of noisy samples in real data, it becomes essential to filter this data. Currently, filtering methods mainly include model-based methods and heuristic-based methods. After filtering, the data needs to be de-duplicated, which can be categorized into two methods: fuzzy deduplication (such as SimHash(*Detecting Near-Duplicates for Web Crawling Proceedings of the 16th International Conference on World Wide Web*, n.d.)) and exact substring matching deduplication.

(2) Model Architecture Design

Typically, when training Large Model, increases in both data and model scale will enhance training results. However, the larger the model, the more training resources it occupies. Therefore, it's essential to employ DL architectures that can compute efficiently when building Large Model. Currently, Transformers and their variants are the most widely used deep learning architectures for building Large Model. Additionally, the size of the model should match the scale of the training data to prevent the waste of computational resources(Hoffmann et al., 2022).

(3) Fine-tuning for Downstream Tasks

After pre-training a Large Model, in order to adapt it to downstream tasks, it requires fine-tuning. This involves introducing specific downstream task data, allowing the Large Model to continue training on top of the pre-trained weights until it meets the requirements of the downstream task. Efficient fine-tuning methods are crucial for maximizing the performance of Large Model. At present, common fine-tuning methods include LoRA(E. J. Hu et al., 2021), Adapter(R. Zhang et al., 2023), Prefix-tuning(X. L. Li & Liang, 2021), P-turning(X. Liu et al., 2021), Prompt-tuning(Lester et al., 2021), and RLHF(Christiano OpenAI et al., 2017), among others.

### 3.1.2 Large Model Advantage Analysis

(1) Generalization Ability

Generalization ability refers to the model's capability to make accurate predictions with new data. When traditional DL models have too many parameters or overly complex structures, they tend to overfit(Philipp & Carbonell, 2018). However, after going through pre-training and fine-tuning, Large Model can capture more details and generalize better to new datasets and tasks. Therefore, compared to traditional DL model, Large Model has a stronger generalization ability.

(2) Emergent Abilities

Traditional DL designs model to solve specific tasks(Ganaie et al., 2022), whereas Large Model uses vast amounts of data for pre-training and then fine-tune for specific downstream tasks, allowing them to adapt to most tasks. As the parameter scale of Large Model increases to a certain threshold, the model's ability to handle certain problems grows rapidly, showing emergent abilities(Wei, Tay, et al., 2022). Therefore, compared to traditional DL models, Large Model can develop unexpected new capabilities during training, demonstrating emergent abilities.

(3) Reasoning Ability

Traditional DL struggles to complete a new task like humans can with just a few examples or instructions, such as in NLP. In contrast, Large Model(Weng et al., 2022), like GPT-3, excel in this aspect, effectively handling reasoning tasks like word replacement in sentences or basic arithmetic. Therefore, Large Model has a stronger reasoning ability.

(4) Decision-making Ability

The decision-making ability is primarily exhibited in decision-intelligence Large Model. Existing decision-intelligence Large Model, like GATO(Reed et al., 2022), transplant pre-trained Large Model into decision-making tasks. They have made progress in reinforcement learning decisions and operations optimization decisions. Compared to traditional decision-making algorithms, Large Model excels in cross-task decision-making and rapid transfer abilities.

(5) Generative Ability

The generative ability of Large Model mainly refers to the quality of content the model can produce and is one of the core capabilities of Large Model. For instance, LLM like ChatGPT can predict and supplement subsequent text content based on previous text; One-2-3-45 can generate 3D models from a single image or even textual content(M. Liu, Xu, et al., 2023). It's evident that Large Model can capture and analyze information better, producing high-quality content. Their performance in generation ability is notably outstanding, offering a significant advantage.

### 3.1.3 Large Model Mature Cases

At present, Large Model can be divided into general field LM and specialized field Large Model. Some mature cases are shown in the following table.

**Table 1 LM part of the mature case**

| Category | Model | Research and development organization | Time | Parameter scale | Pre-training data size | Base model | Field | Open source or not |
|---|---|---|---|---|---|---|---|---|
| general field Large Model | T5 | Google | 2019-10 | 11B | 1T tokens | —— | NLP | Yes |
| | Ernie 3.0 Titan | Baidu | 2021-11 | 260B | 300B tokens | —— | NLP | No |
| | GLM | Tsinghua University | 2022-10 | 130B | 400B tokens | —— | NLP | Yes |
| | BLOOM | HuggingFace | 2022-11 | 176B | 366B tokens | —— | NLP | Yes |
| | LLaMa | Meta | 2023-2 | 65B | 1.4T tokens | —— | NLP | Yes |
| | GPT-4 | OpenAI | 2023-3 | 1800B | 13T tokens | —— | NLP | No |
| | PanGu--Σ | Huawei | 2023-4 | 1085B | 329B tokens | PanGu-α | NLP | No |
| | Guanaco-65B | UW (University of Washington) | 2023-5 | 7B/13B/33B/65B | —— | —— | NLP | Yes |
| | PaLM 2 | Google AI | 2023-5 | 340B | 3.6T tokens | —— | NLP | No |
| | Falcon | TII | 2023-6 | 1B/7B/40B | 1500B tokens | —— | NLP | Yes |
| | MPT-30B | MosaicML | 2023-6 | 30B | 2.2T tokens | —— | NLP | Yes |
| | Claude v2 | Anthropic | 2023-7 | 50B | —— | —— | NLP | No |
| specialized field Large Model | Vicuna-33B | LMSYS | —— | 33B | —— | LLaMa-13B | NLP | Yes |
| | Alphafold | DeepMind | 2022 | —— | —— | —— | Biology | Yes |
| | HuatuoGPT | Cuhk (Shenzhen) | 2023-2 | —— | —— | LLaMA-7B | Healthcare | Yes |
| | BloomBergGPT | Bloomberg | 2023-3 | 50B | 363B tokens | —— | Financial | No |
| | DoctorGLM | —— | 2023-4 | 6B | 1T tokens | chatglm-6b | Healthcare | Yes |
| | TaoLi LLaMa | Beijing Language and Culture University, Tsinghua University, Northeastern University, Beijing Jiaotong University | 2023-6 | 7B | 1.2T tokens | LLaMA | Education | Yes |
| | QiZhenGPT | —— | 2023-6 | —— | —— | LLaMA | Healthcare | Yes |
| | chatglm-maths | Tsinghua University | 2023-6 | 6B | 1T tokens | chatglm-6b | Math | Yes |
| | ChatLaw-33B | Beijing University | 2023-7 | 33B | —— | Anima | Law | Yes |
| | Med-PaLM | DeepMind | 2023-7 | 12B/84B/562B | 3600B tokens | —— | Healthcare | Yes |
| | Medical GPT-zh | —— | —— | —— | —— | chatglm-6b | Biology | Yes |

## 3.2 The Overview of Large Language Model

In the various types of Large Model, the development of Large Language Model (LLM) is particularly notable. It is an NLP model based on DL. By training on large-scale corpora, it learns the probability distribution of words, phrases, and sentences, enabling it to generate fluent and semantically accurate text. Currently, there are application examples on the market including the GPT series(Brown et al., 2020; OpenAI, 2023; Openai et al., n.d.; Radford et al., n.d.) and others.

### 3.2.1 Introduction to LLM

As a technology designed for computers to understand human language, the fundamental problem of NLP is word representation. This means converting words, the most basic linguistic units in natural language, into formats that machines can understand (Fan et al., 2023). However, there are many challenges in the process of(Fan et al., 2023) word representation. To address this, DL introduced word embedding, which constructs a low-dimensional dense vector space from vast amounts of text. This reduces the demand for storage space and addresses the sparsity issue of words.

Based on these techniques, statistical language models were developed. Their primary function is to predict the next word based on the context. The approach used to achieve this is called the N-Gram language model. The N-Gram language model follows the Markov assumption, estimating the likelihood of the next word by counting the frequency of words that appear after a certain number of preceding words.

Building on the foundation of language models, the advent of DL has driven the continuous evolution from neural language models to pre-trained models and up to the current LLM (Casola et al., 2022; Fan et al., 2023; Tripathy et al., 2021).

### 3.2.2 LLM Mainstream Architecture

The Transformer architecture, due to its exceptional parallelism and capacity, has become the base structure backbone for the development of various LLM. This makes it possible to scale language models to hundreds of billions or even trillions of parameters. Generally speaking, existing LLM can be mainly divided into three types: encoder-decoder, causal decoder, and prefix decoder. Below is an introduction to some of the more representative architectures among them:

(1) Seq2Seq(Sequence-to-Sequence)

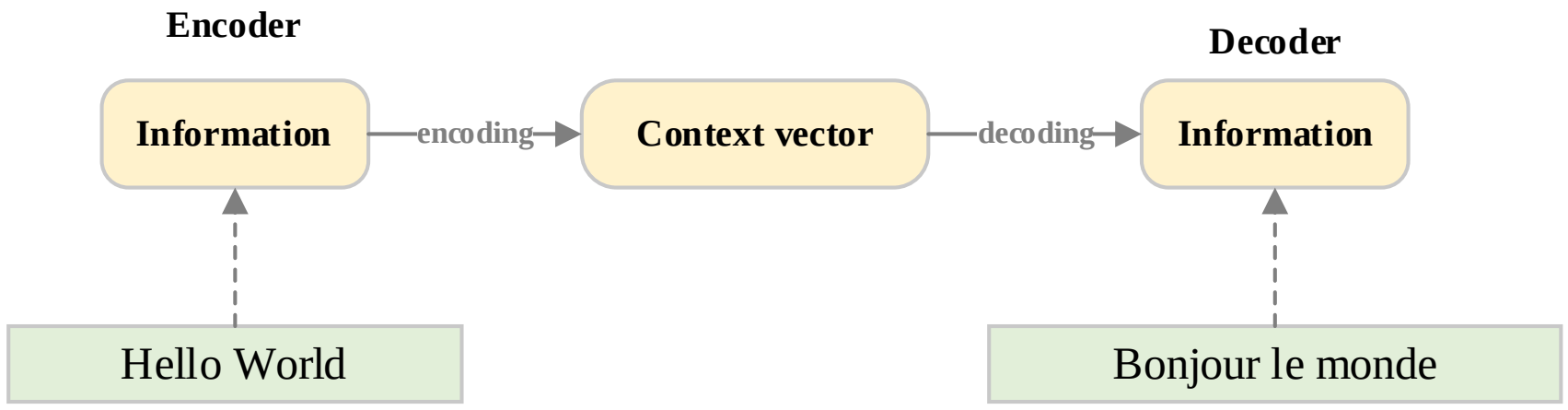

**Fig. 3-2 Seq2Seq Architecture**

Seq2Seq (Sequence-to-Sequence) is a deep learning architecture used for NLP tasks, consisting of an encoder and a decoder (Lewis et al., 2019).

Seq2Seq performs well in tasks like machine translation, text summarization, and dialogue generation. Its strength lies in handling input and output sequences of varying lengths and being able to learn the complex mappings between them. However, the computational complexity of Seq2Seq is relatively high when dealing with long sequences. Its ability to handle rare vocabulary and complex grammatical structures is limited. To solve these problems, researchers have proposed improved Seq2Seq models, such as those which based on the Transformer model use *Self-Attention* mechanisms to better capture dependencies in sequences(Chen et al., 2021; Ouyang et al., n.d.)..

(2) Transformer

The Transformer is a deep learning architecture used for NLP tasks. It evolved from Seq2Seq and has been widely applied to other NLP tasks, such as text classification, named entity recognition, and question-answering systems.

The design philosophy behind the Transformer is based on the Self-Attention， which can weight each position in the input sequence and incorporate these weighted results as part of the inputs, allowing the model to better understand contextual information(Bahdanau et al., 2014)..

The Transformer architecture consists of encoders and decoders(Vaswani et al., 2017). The encoder is responsible for encoding the input sequence into a series of high-dimensional vector representations. The decoder then generates the corresponding output sequence based on the encoder’s outputs and the target sequence. Both the encoder and the decoder are composed of multiple stacked self-attention layers and feed-forward neural network layers.

In addition to the encoder and decoder, the Transformer also introduces position encodings to represent the order information of each position in the input sequence. By encoding the positional information into fixed-length vectors, it can be added to word embeddings to produce the final input representation.

Models that adopt the Transformer architecture can be roughly divided into bidirectional models that only use the encoder, represented by BERT, and unidirectional models that only use stacked decoders for training, represented by GPT(Devlin et al., 2018; Y. Liu et al., 2019; Openai et al., n.d.).

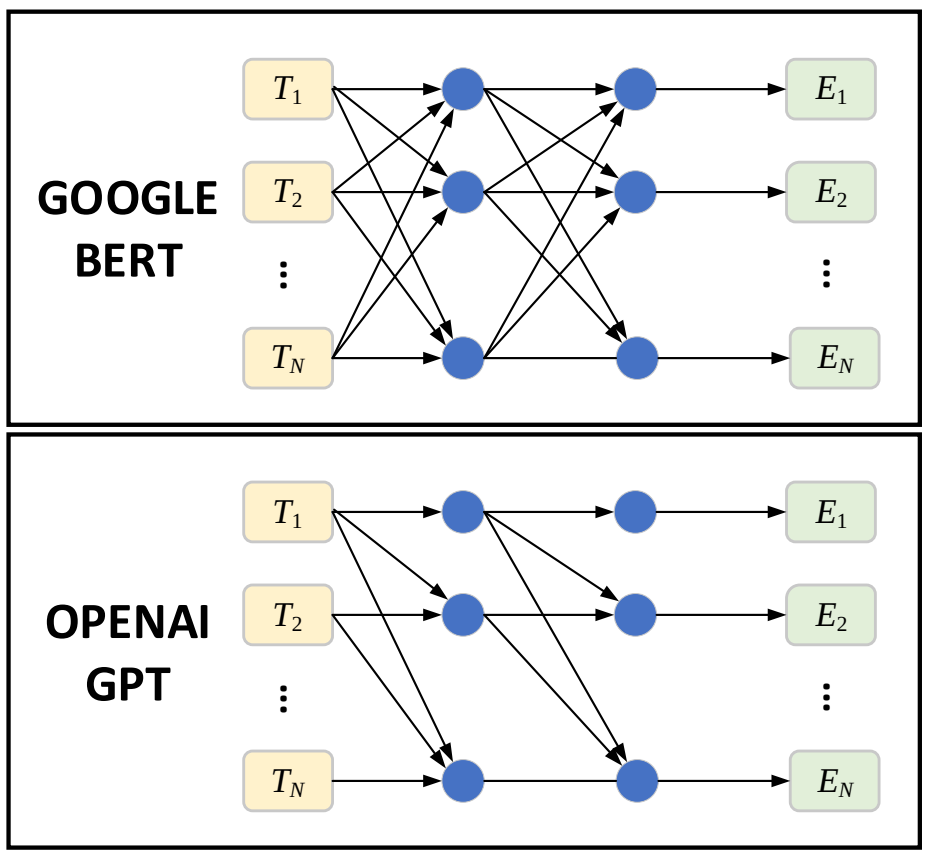

**Fig. 1-3 Two Representatives of the Transformer Architecture**

Overall, through the combination of the Self-Attention and feed-forward neural network layers, the Transformer architecture allows models to better handle NLP tasks. It has achieved significant performance improvements in tasks like machine translation and has become an important benchmark model in the current NLP field.

### 3.2.3 LLM Key Technologies

The technology of LLM is based on existing NLP techniques. The establishment of LLM is achieved through three key technologies: pre-training of the language model, fine-tuning of the LLM, and designing appropriate prompt strategies for the tasks that various LLM will address.

#### 3.2.3.1 Pre-training

Pre-training is one of the critical steps in LLM(W. Khan et al., 2023; Sohail et al., 2023) and serves as the foundation for the powerful capabilities of these models. By utilizing vast amounts of unlabeled text data for unsupervised learning, pre-training can capture the statistical properties and contextual relationships of the language(Ruan & Jin, 2022).

During the pre-training phase, the model is fed with a large amount of text data and trained using an autoregressive generation approach. This means the model predicts the next word or character based on the already generated portion of the text, gradually producing coherent text.

After pre-training, LLM acquire extensive linguistic knowledge and probability distributions. This enables them to generate coherent and natural text, enhancing their understanding and generation capabilities regarding language(Cao et al., 2023).

#### 3.2.3.2 LLM Adaptation Fine-tuning

Fine-tuning of LLM can be described as the process of further training and optimization of the model on supervised annotated data, following its pre-training phase(Chen et al., 2021). This can be subdivided into two primary methods: *instruction tuning* and *alignment tuning*. While instruction tuning aims to augment the capabilities of LLM, alignment tuning seeks to harmonize the model's reactions to textual stimuli with human values, guarding against potential moral dilemmas(Ouyang et al., n.d.).

(1) Instruction Tuning

Instruction tuning is a technique of adjusting the model's inputs. By modulating the instructions or prompts given to the model, one can enhance the quality of the model's generated outputs. In instruction tuning, there is no retraining of the model parameters. Instead, it is achieved by devising more apt input directives, leading the model to produce more accurate and relevant responses. The objective of instruction tuning is to impact the model's outputs by framing more precise directives, ensuring it aligns better with user requirements. This type of fine-tuning can enhance the model's adaptability, accuracy, and generalization capabilities, rendering it effectively operational in instruction-related applications.

(2) Alignment Tuning

Despite the impressive capabilities demonstrated by LLM, there remains the potential to inadvertently harm human society, such as by fabricating misinformation or emitting biased responses. Therefore, it's imperative to align these models with human values, ensuring they uphold standards of utility, honesty, and harmlessness. To this end, some studies have introduced reinforcement learning based on human feedback (RLHF(Christiano OpenAI et al., 2017)). By gathering high-quality human feedback, the learning reward model encourages the LLM to gradually adapt to human standards. However, certain research indicates that alignment tuning might, to some extent, diminish the model's overall versatility.

#### 3.2.3.3 Design appropriate prompting strategies based on the task

After pre-training and fine-tuning, to deploy an LLM effectively, it is essential to devise appropriate prompting strategies for various tasks(Radford et al., n.d.). This has led to the development of In-Context Learning (ICL) and Chain-of-Thought Prompting (COT).

(1) In-Context Learning (ICL)

ICL was first introduced in GPT-3 and subsequently became a standard method for using LLM(W. Khan et al., 2023; Sohail et al., 2023; Zhao et al., 2023)..

The fundamental premise of ICL is that, during text generation, it relies not only on the current input context but also uses previously generated text as an extended context. In every iteration, ICL merges the previously produced text with the current input context, resulting in a richer contextual representation. This iterative approach aids in addressing long-distance dependency challenges, enhancing the quality and coherence of the generated text(Dong et al., 2022).

(2) Chain-of-Thought Prompting (COT)

COT involves the model generating outputs through a series of intermediate steps related to the inputs, presenting possible reasoning pathways or associated concepts(Sohail et al., 2023; Zhao et al., 2023). These prompts elucidate how the model arrives at specific answers or perspectives, shedding light on some of the model's internal mechanisms in reasoning and linguistic expression.

In LLM, COT leverages the model's linguistic comprehension and generation abilities combined with internal representations and attention mechanisms. This method offers insights into understanding the model's reasoning process and provides a more in-depth interpretation of the model's outputs.

Based on the COT, which transforms the mind from a chain to a graph, the ETH Zurich research team proposes the graph of thoughts (GOT) approach to improve the LLM's ability in reasoning and verbal expression.

### 3.2.4 LLM Advantages Analysis

Leveraging vast datasets and sophisticated learning algorithms, LLM exhibit enhanced power, precision, and flexibility in NLP tasks. Surveys of Large Model indicate that numerous industries, including finance, law, and healthcare, are increasingly integrating LLM to construct specialized field applications. Regarding the PHM field, the strengths of LLM can be also harnessed to pioneer new research paradigms:

(1) Information Extraction and Decision-making Support

LLM can automatically extract key information from vast textual data, recognize fault patterns, and offer decision-making support. They are adept at thoroughly analyzing and understanding an abundance of information, such as machinery operational data, repair records, technical manuals, etc., assisting engineers and maintenance personnel in rapidly pinpointing issues and devising solutions.

(2) Constructing Expert Knowledge Bases

LLM can establish a knowledge base saturated with extensive PHM expertise. By learning and comprehending prior fault cases, repair experiences, and other knowledge, these models can present relevant solutions and recommendations. Moreover, they can furnish coherent explanations(Devlin et al., 2018), elevating the accuracy and efficiency of fault diagnosis.

(3) Multi-field Applicability

Exhibiting strong generalization capacities (Brown et al., 2020) and knowledge transfer abilities (Xie et al., 2023), LLM are viable for machinery PHM across varied fields, encompassing industrial machinery, aerospace, ships, vehicles, etc. Their generalization and adaptability enable customization and optimization based on specialized field needs, thereby catering to diverse industry requirements for PHM.

(4) Multi-modal Data

LLM are capable of concurrently processing various data formats, such as text, images, sounds, etc. This implies they can extract information from textual descriptions of system while also analyzing system-related images, sounds, and other data types, culminating in a comprehensive assessment.

(5) Algorithm Intelligent Recommendations

In the context of Q&A, LLM can produce personalized recommendation outcomes based on users' preferences and historical actions(Hou et al., 2023). Hence, they can be deployed for intelligent PHM algorithm recommendations, suggesting algorithms suitable for specific PHM cases to engineers, fulfilling varied task demands.

### 3.2.5 LLM Mature Cases

Major tech companies and academia have delved deep into NLP for years, continually unveiling innovative LLM. By focusing on some typical LLM and understanding their characteristics and application status in the industry, we can provide cases for the combination of PHM and LLM.

(1) GPT-4(OpenAI, 2023)

The latest generative pre-trained transformer model in the GPT series, GPT-4 is released by OpenAI. Supporting multimodal format inputs, GPT-4 is adept at resolving complex issues across disciplines like mathematics, coding, and healthcare, even without explicit prompts. GPT-4 has not yet released technical details and cannot be deployed on this basis, but its model performance is still leading the industry, and some Large Model in specific fields will generate training datasets based on the results of conversations with GPT-4.

(2) Llama(Touvron et al., 2023)

Llama, a collection of foundational language models of four different parameter scales, is released by Meta AI. During the fine-tuning phase, Llama undergoes supervised training based on labels specific to a given task, further enhancing the model's performance. At present, many Large Model in specific fields are choosing Llama as the base model to meet the needs of different downstream industries by adding different training sets of specialized fields. The latest Llama 2 has also been open source, compared with the previous generation in performance, reasoning efficiency has been significantly improved.

## 3.3 Summary

In summary, Large Model demonstrates outstanding capabilities in content generation, generalization, reasoning, and decision-making. The success of these models further validates the synergy of vast data, high computational power, and efficient algorithm architecture. The development process of Large Model harnesses an array of artificial intelligence-supported technologies, such as machine learning, computer vision, knowledge graphs, and natural language understanding. This showcases the pivotal role AI technologies play in the construction of Large Model.

The reasoning, decision-making, generalization, emergent, and generative attributes inherent in Large Model offer significant opportunities to address the current challenges in PHM, such as weak algorithm generalization, underutilization of field knowledge, and poor algorithm & model versatility. This convergence presents an organic juncture for the integration of Large Model technologies with PHM techniques. Furthermore, contemporary PHM extensively employs AI technologies to tackle related issues, such as fault diagnosis. Consequently, there exists a technical foundation and feasibility for the union of PHM with Large Model technology at a technological level.

Blending the techniques related to Large Model with PHM can address current PHM challenges regarding algorithm, field knowledge, generalization performance, and developmental thresholds. This fusion promise to steer PHM towards a more efficient research paradigm.

# 4. Concept of PHM-LM and Progressive Research of Paradigms

As research into Large Model burgeons at an accelerated pace, the technical merits of generalization capacity, emergence capacity, and generative capacity brought forth by these models offer significant opportunities for the progressive evolution of the PHM. Consequently, when tackling the challenges and predicaments encountered in PHM, the comprehensive utilization of Large Model's capacities and their underlying principles introduces novel perspectives and theories for the future development of PHM(Jia et al., 2023; Jiao et al., 2020; Lei et al., 2020). Building upon an analysis of the developmental challenges in the PHM and the distinctive advantages offered by Large Model, this chapter introduces, for the first time, the concept and essence of PHM-LM. Furthermore, based on the form and degree of the integration between Large Model and PHM, it progressively outlines 3 typical paradigms of innovative research for the advancement of PHM-LM exploration and application, ranging from fundamental to advanced. Subsequently, under 3 typical innovation paradigms, considering the characteristics and capabilities of each paradigm, we propose the key PHM capabilities that can be enhanced by Large Model in different paradigms, as well as feasible technical approaches to enhance these capabilities. This section offers researchers engaged in the integration of Large Model with PHM a comprehensive and progressively detailed technical study roadmap and specific solution strategies.

## 4.1 Concept and Connotation of PHM-LM

Generally, a PHM-LM refers to an artificial intelligence model characterized by its vast parameter scale and complexity, deeply integrated with the specificities of the PHM specialized field, and capable of serving the entire life cycle of product PHM. The design and application of these PHM-LM aim to address the myriad complex tasks in PHM in a more intelligent, precise, and general manner. Given the multifaceted nature of PHM, these Large Model can be further categorized into:

(1) Object-oriented PHM-LM, such as models for bearing PHM, gear PHM, motor PHM, and electronic product PHM;

(2) Task-oriented PHM-LM, like models for data generation, solution generation, and verification & validation;

(3) Algorithm & model-oriented PHM-LM, including models for fault diagnosis, fault prediction, and maintenance decision-making.

Compared to traditional PHM systems, PHM-LM exhibit significant advancements in aspects like data, knowledge, model architecture, and capabilities. Traditional PHM models tend to be narrow in scope, handling limited data and knowledge. In contrast, the PHM-LM offer robust capabilities in processing multi-modal data, effectively handling both complex and non-ideal data conditions. On another front, while traditional PHM models often serve as small, purpose-specific models, the PHM-LM is the synergistic outcome of integrating Large Model technology with PHM modeling techniques. Essentially, Large Model are the consolidation and fusion of these smaller models. In terms of capabilities, the PHM-LM facilitate the resolution of a myriad of complex tasks throughout the entire PHM life cycle in an intelligent, precise, and general manner. They prove invaluable to PHM designers, positioning themselves as top-tier intelligent experts in the PHM field.

The PHM-LM, as an advanced generative AI model, is equipped with multi-modal data-driven capabilities, superior generalization, and practical performance. These attributes are poised to spur innovative breakthroughs in technical research within the PHM field and catalyze a comprehensive overhaul of its service models. In the sections that follow, we will delve into a step-by-step exposition and analysis of three representative innovative research paradigms for advancing the study and application of PHM-LM. Specifically, these paradigms are:

- Paradigm I: PHM Paradigm based on LLM
- Paradigm II: Parallel Paradigm of LLM and PHM Model
- Paradigm III: Construction and Application Paradigm of PHM-LM

## 4.2 Paradigm I: PHM Paradigm based on LLM

In the context of customized requirements, extensive research on fault prognostics(Boškoski et al., 2015), fault diagnosis(J. Liu, 2022; J. Sun et al., 2022), and health management(S. Khan & Yairi, 2018) has been conducted in PHM(Meng & Li, 2019). However, due to challenges such as insufficient generalization, there remains significant room for optimizing the development of PHM practices. Current obstacles, such as a lack of efficient knowledge integration(Gay et al., 2021), repetitiveness in textual design, complexity in data modalities(Farsi & Zio, 2019), and the difficulty of algorithm & model development(Vogl et al., 2019), have resulted in inefficiencies, poor generalization, and limited interpretability in traditional PHM services such as knowledge retrieval, Q&A, explanatory supplementations, redundant textual tasks, algorithm & model-assisted development, knowledge-based diagnostics, and maintenance decision-making. Generative LLM, with their strong generalization, reasoning, and comprehension capabilities, have already revolutionized sectors like finance(S. Wu et al., 2023), healthcare(Thirunavukarasu et al., 2023), and law. By offering rapid information retrieval, text processing, and case analysis, these models have enhanced knowledge utilization, work efficiency, and generalization in their respective fields. Such successes are testimonies to the potential improvements that LLM technology can bring to specialized fields of research.

Given this, a PHM paradigm rooted in LLM will exploit the language reasoning and generalization capabilities of foundational LLM. Through rapid fine-tuning with field-specific knowledge from the PHM field, this paradigm aims to build a specialized field LLM tailored for typical scenarios within the PHM realm, thereby addressing a host of issues present in this field. Anchored in this concept, the Paradigm I designs applications such as PHM knowledge engineering, LLM diagnostic expert, assisted algorithm & model development for PHM, text generation for PHM, and LLM-driven maintenance decision-making. The ultimate goal is to holistically enhance traditional PHM services, including knowledge integration, expert diagnostics, algorithm & model development, textual solution design, and maintenance decision-making, leveraging next-generation artificial intelligence LLM technology to elevate efficiency, generalization, and decision-making capabilities in PHM.

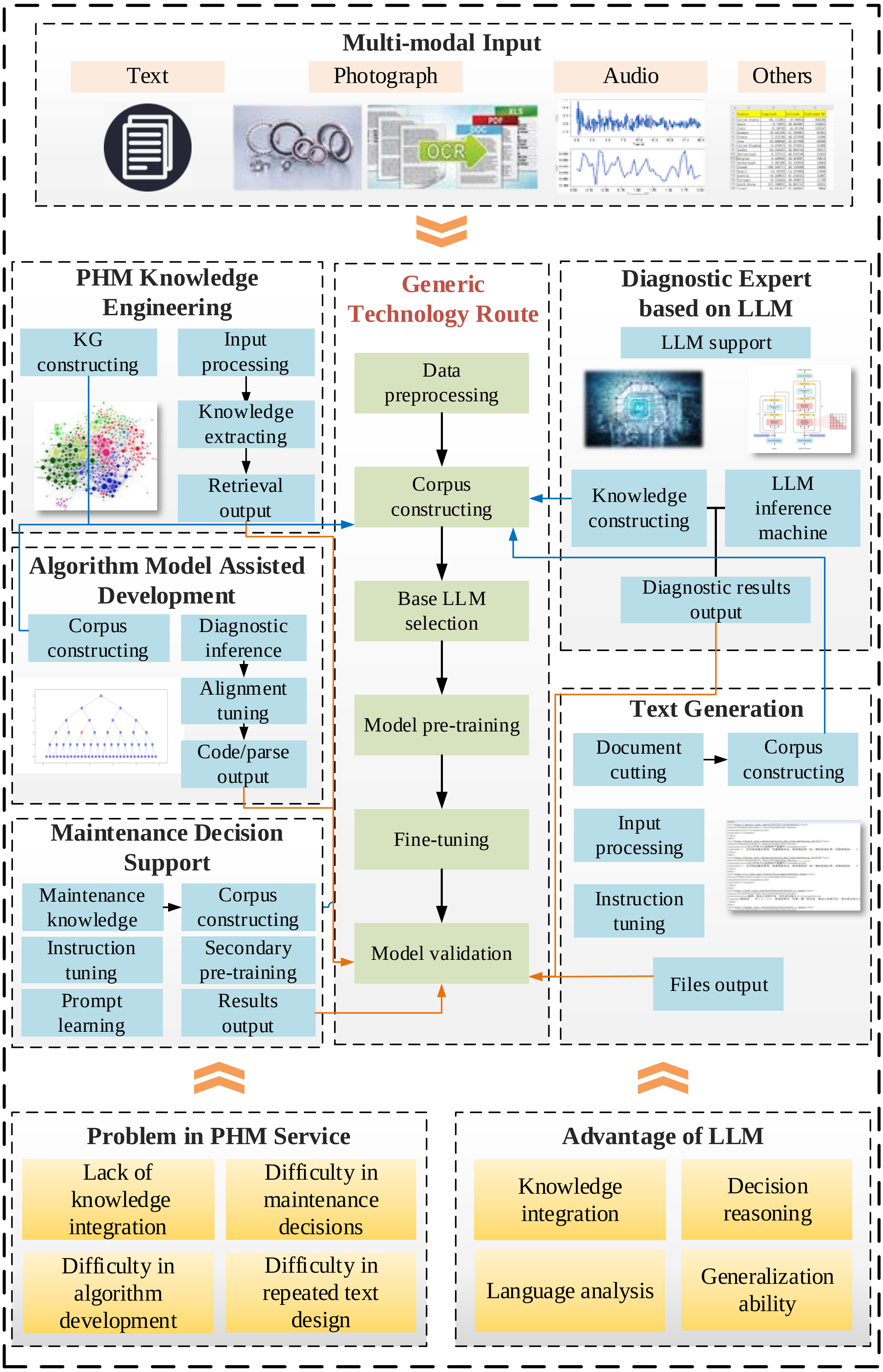

**Fig. 4-1 PHM Paradigm based on LLM**

### 4.2.1 Approach 1: Knowledge Engineering of PHM based on LLM and Knowledge Graph

**Name: Knowledge Engineering of PHM Based on LLM and Knowledge Graph**
**Problems: Poor utilization of knowledge, low retrieval efficiency**
**Advantages of LM: Linguistic analysis ability, knowledge reasoning ability**
**Inputs: Retrieval needs**
**Outputs: Knowledge retrieval results, reasoning results**

**Typical application forms: knowledge retrieval, interactive Q&A, intelligent reasoning**

**Fig. 4-2 Information Card of Knowledge Engineering of PHM based on LLM and Knowledge Graph**

Based on the fusion of LLM and knowledge graphs, the PHM knowledge engineering aims to address the challenges of low knowledge utilization, inefficient retrieval, and insufficient decision interpretability. Guided by the principle presented in Paradigm I, which focuses on the application of LLM in specialized fields to tackle a range of issues in PHM, this approach leverages the strengths of LLM in language analysis, knowledge reasoning, and generalization. This approach revolves around the task of taking retrievable content as input and producing knowledge retrieval, reasoning, and decision outcomes. It is applied to various scenarios within PHM, including knowledge retrieval, interactive question answering, and reasoning-based decision-making. By employing the capabilities of LLM and knowledge graphs, this approach seeks to achieve efficient, accurate, and reliable knowledge querying and datasets interaction analysis, thereby contributing to the advancement of the PHM field(Y.-F. Li et al., n.d.).

LLM possess remarkable capabilities in language analysis, knowledge reasoning, and generalization(Zhao et al., 2023). Moreover, architectures such as Transformers and RetNet offer parallel processing and enhanced computational efficiency. Their ability for context learning and fine-tuning empowers Transformer-based models to excel in handling vast amounts of information and solving complex problems, rendering them superior in certain fields(Bagal et al., 2022; Y. Xu et al., 2022). However, current LLM still face challenges such as poor real-time data processing and adapting to emerging trends. Without a comprehensive and real-time knowledge base, these limitations hinder their application in PHM. Leveraging the advantages of knowledge graphs in data mining, knowledge integration, and information processing, knowledge engineering in PHM based on knowledge graphs has achieved intelligent fault diagnosis, fault reasoning, and fault localization for specific entities(X. Tang et al., 2023). Nevertheless, relying solely on knowledge graph technology can only accomplish limited-scale PHM knowledge engineering for small-scale entities, falling short in enhancing the generalization capabilities of PHM models. Therefore, combining the strengths of LLM and knowledge graphs, this approach proposes a PHM knowledge engineering framework based on both components. This integrated approach aims to address the drawbacks of poor real-time performance and knowledge scarcity in LLM, as well as the limitations of knowledge graph generalization and limited scale. The result is a solution that enables rapid, credible, wide-ranging, and multi-modal data-supported knowledge retrieval, question answering, and decision reasoning functions.

The PHM knowledge engineering based on the interaction between LLM and knowledge graphs comprises two main components. The construction of a PHM knowledge graph involves the consolidation of sources for PHM knowledge. Driven by specific requirements, PHM knowledge is gathered, followed by data cleaning, vectorization, and structured transformation. The next step involves storing this information in a document dataset and constructing a knowledge graph that encompasses relevant concepts and service aspects. This knowledge graph serves as the foundation for data extraction in the interaction with LLM(Y. Tang et al., 2023). In the context of LLM interaction, the system receives various multi-modal inputs related to PHM from users. It deconstructs the user's requirements and processes the questions through vectorization. An intent generalization module enhances the generalization capability of the LLM, and the generalized requirements are used for searching the knowledge graph data(Y. Liu et al., 2022). The system then returns relevant graph concepts and attributes. The output of user queries is generated using the NLP module of the LLM. This process incorporates the reasoning capability of the LLM to validate and evaluate search results. Through this approach, it addresses challenges such as poor generalization in PHM and low knowledge utilization efficiency, achieving efficient, accurate, and trustworthy knowledge retrieval and dataset interaction analysis. This mechanism provides robust technical support for the management of health knowledge related to various devices and their critical components, as well as for PHM decision-making.

#### 4.2.2 Approach 2：Diagnostic Expert based on LLM

**Name:** **Diagnostic Expert Based on LLM**
**Problems:** **Low troubleshooting efficiency**
**Advantages of LM:** **Knowledge reasoning ability, decision-making recommendation ability**
**Inputs:** **Fault descriptions, audio signals, pictures, etc.**
**Outputs:** **Fault diagnosis reasoning results**

**Typical application forms: expert experience-based fault diagnosis**

**Fig. 4-3 Information Card of Diagnostic Expert based on LLM**

Diagnostic expert based on LLM has faced limitations that have led to the stagnation of expert systems' knowledge bases and reasoning engines in the field of diagnostic expert systems. In order to overcome these limitations, the concept of using LLM to address a range of issues in the PHM field, as discussed in Paradigm I, is extended to provide a solution. Leveraging the strengths of LLM in knowledge reasoning and decision recommendation, a new approach is introduced. This approach involves utilizing the knowledge reasoning and decision recommendation capabilities of LLM to address the limitations of knowledge-based diagnostic expert systems. It focuses on incorporating various inputs such as textual descriptions of faults, audio signals like vibration patterns, images of faulty system, and more. These inputs are processed by the LLM, which then generates outputs related to potential directions for fault diagnosis research and diagnostic results that can be referenced. The application of this approach, in forms such as diagnostic result reasoning, contributes to the advancement of PHM technology by enhancing the efficiency of fault diagnosis.

The development of expert systems based on knowledge-based diagnosis has reached a bottleneck due to its inherent limitations(Akram et al., 2014; Cowan, 2001). However, the knowledge reasoning and decision recommendation capabilities of LLM have opened up new avenues for the previously stagnant expert systems. Currently, in fields like healthcare(Meskó & Topol, 2023) and finance, LLM utilizing expert knowledge have demonstrated enhanced diagnostic capabilities through knowledge-based reasoning. This emphasizes the potential of LLM to improve diagnostic systems within specialized fields. Thus, leveraging the strengths of LLM, a novel approach is proposed: the creation of a generative LLM diagnostic expert system formed by the integration of knowledge repositories and reasoning engines.

Users are required to provide a substantial amount of expertise in PHM to construct the knowledge repository, thereby endowing the expert system with proficiency in fault diagnosis. The diagnostic expert system based on the LLM necessitates a signal processing module for handling multi-modal inputs such as images of faulty components, FMECA tables, videos, etc., and transforming them into a format compatible with the language-processing capabilities of the LLM. The reasoning engine needs to leverage the robust reasoning capabilities of the LLM. It accomplishes this through context learning, fine-tuning, and similar techniques. This enables the system to analyze fault patterns within the input data, cross-reference them with the knowledge and data stored in the corpus, and generate logically coherent and convincing diagnostic outcomes. Additionally, the reasoning abilities of the LLM are employed to evaluate whether the results meet the requirements of the diagnosis, thereby catering to user needs.

### 4.2.3 Approach 3：PHM Algorithm & model Assisted Development

**Name:** **PHM Algorithm Modeling Assisted Development**

**Problems:** **High requirements for specialized knowledge and code ability**

**Advantages of LM:** **Code generation and analysis capability**

**Inputs:** **Required code function, code to be analyzed**

**Outputs:** **Required code, code function analysis**

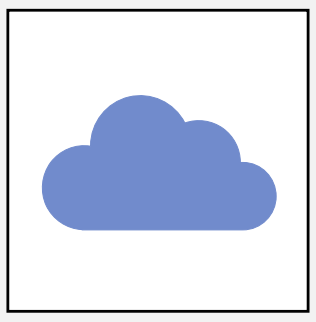

**Typical application forms: partial code completion, code function analysis, etc.**

**Fig. 4-4 Information Card of PHM Algorithm & model Assisted Development**

The PHM algorithm & model assisted development aims to address challenges related to the high demand for specialized knowledge and coding proficiency in the development process. Guided by the principle of the Paradigm I, which leverages the application of LLM in specialized fields to address a series of challenges in the field of PHM, this approach capitalizes on the advantages of the LLM's code generation and analysis capabilities. This approach aims to facilitate PHM algorithm & model development tasks by using natural language descriptions of code requirements and inputting code segments that need explanation. The outputs encompass the desired code, functions, models, and descriptive analyses of model functionalities. This assists in tasks such as code functionality analysis and code completion, resulting in lowered barriers to algorithm & model development and increased efficiency in the development process. Through these typical applications, the development of PHM technology is enhanced, leading to reduced complexity in algorithm & model development and improved efficiency.

In order to develop high-performing PHM algorithms & models, it requires individuals with a strong background in specialized knowledge and coding proficiency. Current LLM have already demonstrated the capability to perform basic code writing(Nijkamp et al., 2022) and interpretation (Leinonen et al., 2023). They hold significant potential in assisting algorithm & model development. However, while LLM possess general code writing and interpretation capabilities, they may not meet the professional standards and requirements for algorithm development on a large scale using collected data. To mitigate the need for high algorithm development skills among PHM algorithm & model developers(Rask Nielsen & Holten Møller, 2022), it is possible to entirely leverage the advantages of LLM in functions such as code writing and analysis. By utilizing LLM for tasks like code completion and code functionality analysis in the realm of PHM algorithms, the difficulty for users in developing PHM models can be reduced, leading to increased efficiency in the development process.

To construct a corpus for PHM algorithms & models and their associated code based on historical data, the aim is to facilitate the pre-training of LLM. This pre-training equips the models with the capacity to engage in high-quality PHM model knowledge reasoning and model generation. Subsequently, utilizing fine-tuning methods, the LLM can analyze existing models and automatically offer developers completion code containing functional analysis explanations for users to select. The models can also detect and highlight warnings or errors present in the code, significantly reducing the complexity of PHM model development(Yuan et al., 2023). Simultaneously, within this framework, a module for handling multi-modal inputs is required. This module receives user descriptions of their code requirements and returns the needed code, functions, models, etc. in the form of answers to users' queries. By following the outlined approach, the objective of reducing the complexity of PHM model development and enhancing the efficiency of the development process can be achieved.

### 4.2.4 Approach 4：PHM Text Generation

**Name:** **PHM Text Generation**

**Problems:** **Much repetitive work, low design efficiency**

**Advantages of LM:** **Linguistic analysis ability, intellectual reasoning ability**

**Inputs:** **Required program type, content and other requirements**

**Outputs:** **Documents that meet the requirements and have clear logic.**

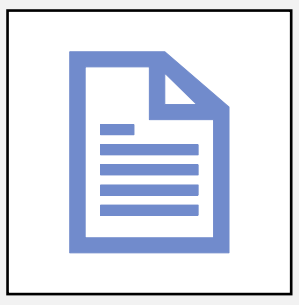

**Typical application forms: text program design, test outline generation, etc.**

**Fig. 4-5 Information Card of PHM Text Generation**

The PHM text generation approach aims to address the challenges of high repetition and low design efficiency in the PHM field. Guided by the principle established in Paradigm I, which leverages the application of LLM in specialized fields, this approach capitalizes on the strengths of LLM, including language analysis, knowledge reasoning, decision-making, and generalization capabilities. By taking elements such as the requirements for generating solutions as input, the approach outputs PHM textual solution documents that adhere to the specified criteria and exhibit clear logic. This task of generating PHM text is exemplified through applications such as designing text and generating test outlines. Through this approach, the development of PHM technology can bring about a reduction in repetitive tasks and an enhancement in the efficiency of textual design, while benefiting from the language analysis, knowledge reasoning, solution decision-making, and generalization capabilities of LLM.

In the practical application of PHM, the diversity of PHM scenarios results in a lack of generalization in PHM solutions. This leads to the necessity of designing custom PHM solutions for different systems and components, resulting in excessive repetition of designs and significant customization demands. This, in turn, diminishes the efficiency of PHM design. Some existing LLM, such as ChatGPT, Llama, GLM, have already demonstrated preliminary capabilities in generating textual solutions(Hirosawa et al., 2023). Additionally, in other fields like finance, healthcare, and law, LLM have shown potential in assisting with professional tasks, achieving accuracy rates as high as 93.3%(J. Liu, Wang, et al., 2023). This showcases the considerable potential of LLM in generating specialized solutions in specialized fields. Thus, it is possible to harness the advantages of LLM in language analysis, knowledge reasoning, decision-making, and generalization capabilities to design a generative PHM textual solution design model. This model can leverage multi-modal inputs of PHM design related requirements to generate the needed PHM design. By doing so, it seeks to address the issue of excessive repetition in current PHM design and enhances the efficiency of PHM design.

Users leverage historical PHM knowledge, solution-related documents, and other corpora to establish a foundation. They employ methods like document segmentation and data classifiers for cleansing, filtering, and structuring the data in preparatory steps, resulting in a pre-training corpus suitable for generating textual solutions. Further, the pre-training of a mature LLM is conducted, imbuing it with robust generalization, reasoning, and multi-modal data processing capabilities. To enhance the effectiveness of the LLM in generating textual solutions, techniques like prompt engineering are employed. By combining usage scenarios and generation objectives, appropriate prompts are designed to guide the LLM's output. For instance, different types of solution requirements, such as diagnostic plans, predictive plans, and design plans, are reflected in the input. Instruction tuning and alignment tuning are employed to adhere to field-specific conventions and to shape the output of the LLM, ensuring it generates PHM textual solutions that meet the requirements coherently. The LLM evaluates the generated results, using specific metrics to measure if the outcomes align with the input requirements. Based on this technical framework, the PHM solution generation assisted by LLM facilitates PHM professionals in their system design tasks. This approach reduces redundancy, enhances efficiency, and improves the effectiveness of PHM text design in the PHM field.

### 4.2.5 Approach 5：Maintenance Decision Support based on LLM

**Name:** **Maintenance Decision Support based on LLM**

**Problems:** **Poor utilization of maintenance knowledge, low decision-making efficiency**

**Advantages of LM:** **Knowledge integration ability, decision-making reasoning ability**

**Inputs:** **Knowledge to be retrieved, fault descriptions, etc.**

**Outputs:** **Retrieved knowledge, maintenance decision making**

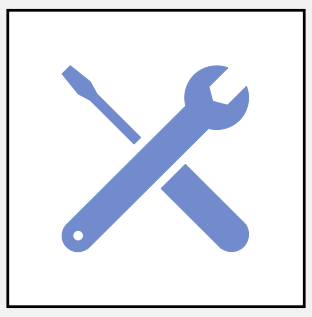

**Typical application forms: maintenance knowledge retrieval, maintenance decision recommendation**

**Fig. 4-6 Information Card of Maintenance Decision Support based on LLM**

The route of maintenance decision support based on LLM aims to address issues like inadequate integration and utilization of maintenance knowledge, as well as low efficiency in maintenance retrieval and decision-making. Guided by the concept introduced in Paradigm I, which utilizes LLM in specialized fields to address a range of problems in PHM, this approach capitalizes on the knowledge integration and decision reasoning capabilities of LLM. Structured data such as FMECA tables and unstructured data like fault descriptions are used as inputs to generate outputs of retrieved knowledge or referenceable maintenance decisions. This process forms the basis of maintenance decision support. By employing the capabilities of LLM, tasks such as retrieval of maintenance knowledge and generation of maintenance decisions are accomplished. This approach seeks to address the challenges arising from insufficient competencies among maintenance personnel, leading to low utilization of maintenance knowledge and poor retrieval efficiency. Through typical applications such as maintenance knowledge retrieval and decision generation, the approach improves the utilization of maintenance knowledge and enhances decision-making efficiency in PHM technology development, thereby resolving challenges stemming from inadequate skills among maintenance personnel and boosting both the retrieval efficiency and utilization of maintenance knowledge.

Maintenance personnel responsible for system repairs require sufficient knowledge and experience. However, due to the complexities arising from the nature of the systems, fault patterns, and maintenance strategies, these personnel often lack the relevant knowledge and experience(Scott et al., 2022). Simultaneously, the underutilization of resources like maintenance manuals and historical maintenance experiences contributes to the inefficiency in knowledge retrieval and generation of maintenance plans, requiring significant time investments and resulting in low maintenance efficiency(Gawde et al., 2023). The knowledge integration and decision reasoning capabilities inherent in LLM offer a potential solution to break through the current bottlenecks in maintenance operations. This notion is reinforced by the successful application of LLM in fields such as finance and healthcare, where they have effectively demonstrated their capacity for knowledge synthesis and reasoning. Consequently, leveraging LLM to achieve generative maintenance knowledge retrieval and decision recommendation can address the challenges faced by maintenance personnel in terms of inadequate knowledge and experience, as well as the inefficiency in knowledge retrieval. This approach aims to enhance the utilization of maintenance knowledge and improve decision-making efficiency in the context of maintenance operations.

To establish a foundation for maintenance decision support using LLM, a series of steps can be taken. Initially, common knowledge from the maintenance manuals and related resources is collected, followed by preprocessing the data to construct a pre-training corpus specific to the maintenance field. This corpus is then utilized to pre-train the LLM. To enhance the model's performance on specific tasks within the maintenance field, a secondary pre-training phase can be undertaken. During this phase, private user data pertinent to maintenance, such as historical repair records and machinery knowledge, is used for secondary pre-training of the LLM. Additionally, the model undergoes fine-tuning with user-specific maintenance data, such as historical maintenance records and machinery knowledge, to further align it with the user's field-specific requirements. The model should be equipped to handle both structured data like FMECA tables and unstructured inputs like fault descriptions and knowledge retrieval queries. Instructions for generating relevant maintenance decisions or retrieving knowledge can be selected through techniques like manual labeling or model-based approaches. The model can then be fine-tuned using supervised fine-tuning (SFT) or instruction tuning to ensure that it effectively aligns with user requirements. Once fine-tuned, the model can generate maintenance decisions or knowledge retrieval results that match user needs. These outputs can be validated and evaluated for compliance with user requirements. In summary, this approach harnesses the robust generative, knowledge utilization, and reasoning capabilities of LLM to provide efficient maintenance decision support. By doing so, the approach seeks to address challenges such as low utilization of maintenance knowledge and the high expertise required from maintenance personnel, leading to improved maintenance efficiency.

## 4.3 Paradigm II: Parallel Paradigm of LLM and PHM Model

Different from the corpus data typically utilized in existing NLP field, data in PHM field commonly comprises time-series data collected by multiple types of sensors, encompassing vibration(Tiwari & Upadhyay, 2021), sound(Ding et al., 2023), electrical signal(Q. Yu et al., 2023), temperature(Jung et al., 2023), pressure(S. Tang et al., 2022), etc. The processing and analysis of typical signal data represent the strengths of conventional PHM models, and extensive research efforts have been dedicated to aspect(Lv et al., 2022; L. Tang et al., 2023). However, conventional PHM models often struggle to meet expected PHM requirements in the case of insufficient data conditions(Pan et al., 2022). In such cases, field-specific knowledge supplementation(Z. Wang et al., 2023; J. Yu & Liu, 2020) or external interventions(J. Wang et al., 2023; T. Zhang et al., 2023) are required to enhance the accuracy and capability of conventional PHM models and their operational patterns. LLM, which is equipped with such capabilities, can harness NLP techniques to process and comprehend specialized field knowledge. Consequently, LLM can be combined with the PHM model to build new parallel models to improve the latter's capabilities, cater to more elevated and detailed PHM demands. This concept of multi-models to realize parallel development has been explored in other specialized fields (Palo et al., 2023). Hence, Paradigm II considers building a parallel development framework of LLM and PHM model upon the mature LLM by integrating its knowledge process, logical reasoning, decision-making, and generalization advantages, which is tailored to guide the training and reinforcement learning processes of PHM models, to solve problems and address challenges related to PHM field. By combining the capabilities of LLM and PHM model through parallel development, we aim to optimize aspects such as algorithm recommendation and updating, thereby enhancing the competency in areas like knowledge-data fusion, interpretability, and credibility. This pursuit culminates in building a novel parallel model designed to tackle the present shortcomings within PHM field, including deficiencies in knowledge utilization, targeted guidance, assistance, and algorithm generalization. The parallel paradigm framework of LLM and PHM model is depicted in Fig.4-7.

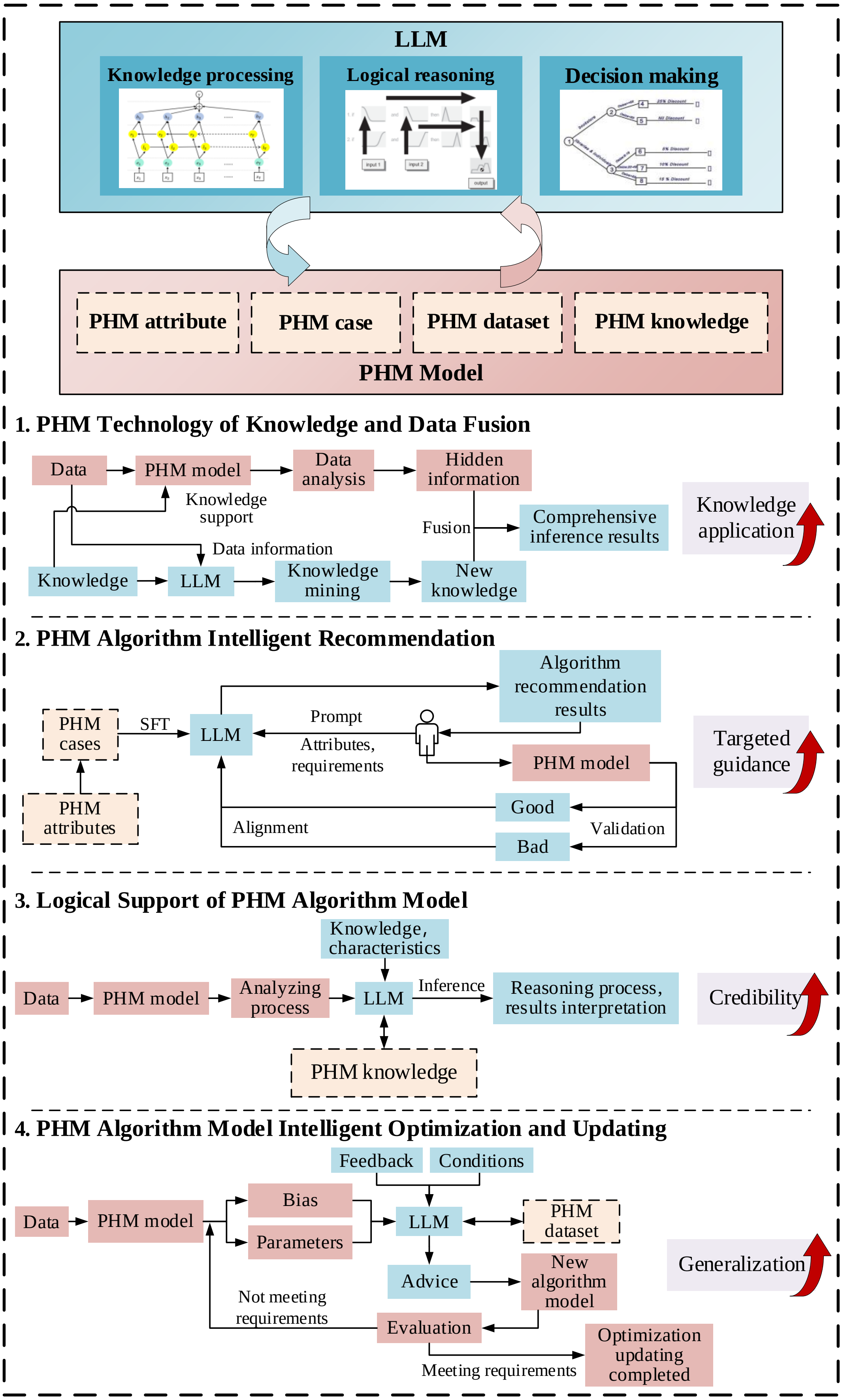

**Fig. 4-7 Parallel Paradigm Framework of LLM and PHM Model**

### 4.3.1 Approach 1: PHM Technology of Knowledge and Data Fusion

**Name:** **PHM Technology for Knowledge and Data Fusion**

**Problems:** **Inadequate utilization of knowledge, poor data-knowledge fusion**

**Advantages of LM:** **Knowledge migration, generalization ability**

**Inputs:** **PHM domain data, knowledge**

**Outputs:** **Synthesized inference results**

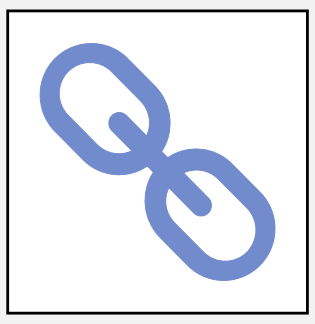

**Typical application forms: knowledge-assisted data analysis, knowledge mining based on data information**

**Fig. 4-8 Information Card of PHM Technology of Knowledge and Data Fusion**

The knowledge and data fusion PHM technology route aims to address the issues of underutilized existing knowledge and inadequate fusion of data and knowledge. Guided by the idea of Paradigm II that LLM lead PHM model to achieve parallel development, this approach leverages the substantial capabilities of LLM in knowledge transferring and generalization. It seeks to accomplish the task of knowledge and data fusion by taking PHM data and knowledge as inputs and producing comprehensive reasoning results. This endeavor is realized through typical applications such as knowledge-assisted data analysis and knowledge extraction based on data information. The ultimate goal is to achieve a thorough integration and utilization of both knowledge and data in the field of PHM, utilizing the knowledge transferring and generalization capabilities of LLM to enhance this fusion process.

The limitations arising from incomplete data acquisition, inconsistent data quality, challenging feature extraction, and insufficient analysis make it difficult for data-driven PHM models to achieve desired outcomes outcomes(Zio, 2022). These challenges necessitate the involvement of relevant field knowledge to facilitate analysis. In contrast, LLM is trained on vast amounts of text, allowing it to accumulate extensive knowledge and information. The synergy between these two models is manifested as follows: the PHM model processes and analyzes the data to uncover hidden patterns, correlations, and regularities, which are then handed over to the LLM for further processing. Simultaneously, the accumulated knowledge and expertise assist in the data analysis process of PHM model, enhancing efficiency and accuracy. The ultimate outcomes are comprehensive results that integrate both data and knowledge. By harnessing the robust knowledge processing capabilities of LLM and the data processing capabilities of PHM model, it becomes feasible to effectively address the shortcomings of data-driven analysis in terms of comprehensiveness and knowledge integration.

Knowledge and data fusion technology involves two main components: LLM and PHM model. The fusion aims to address the shortcomings of PHM model, such as excessive reliance on high-quality data and poor interpretability. Leveraging the data processing capabilities of PHM model, series of data processing algorithms are employed to extract valuable information from raw data, thereby enhancing the accuracy of various model outcomes. Simultaneously, the powerful NLP capabilities of LLM are utilized to comprehend and analyze text, enabling the transferring and generalization of accumulated knowledge. Such knowledge is then fused with the results obtained from data processing, aiding the data analysis performed by PHM model and end up with comprehensive outcomes that combine data support with field expertise. Furthermore, the data supported by PHM model can be exploited by LLM to unearth information from the data, thereby enriching the knowledge base. This approach of knowledge and data fusion not only enhances the generalizability of PHM model, remedies the limitations of solely data-driven approaches, but also validates and enriches the knowledge accumulated by LLM.

### 4.3.2 Approach 2：PHM Algorithm Intelligent Recommendation

**Name:** **PHM Algorithm Intelligent Recommendation**
**Problems:** **PHM algorithm is difficult to choose and design**
**Advantages of LM:** **Strong reasoning ability, knowledge internalization and absorption ability**
**Inputs:** **Object equipment information, algorithm performance elements**
**Outputs:** **PHM algorithm and related configuration**

**Typical application forms: fault diagnosis algorithm recommendation, fault prediction algorithm recommendation, etc.**

**Fig. 4-9 Information Card of PHM Algorithm Intelligent Recommendation**

The PHM algorithm intelligent recommendation approach is devised to address the intricate challenges in PHM algorithm design arising from the diverse and complex elements involved. Guided by the idea of Paradigm II that LLM lead PHM model to achieve parallel development, this approach capitalizes on the knowledge internalization and absorption capability of LLM. This approach takes various elements into consideration, including PHM task and algorithm functional requirements, PHM algorithm background engineering conditions, and PHM algorithm performance elements, among others, as inputs. The outputs are the desired PHM algorithm and related configuration elements. Through typical applications like fault diagnosis algorithm recommendations, fault detection algorithm recommendations, and fault prediction algorithm recommendations, this approach accelerates the generation of PHM capabilities, mitigates the demanding requirements faced by PHM engineers and aids in the development of PHM systems while expediting the design process.

PHM encompasses various core functionalities such as fault diagnosis, fault prediction, among others, which involve a diverse array of algorithms. These algorithms demonstrate robust data processing capabilities but also exhibit performance variations under different data conditions(Z. Xu & Saleh, 2021). Consequently, in practical engineering applications, the choice of algorithms varies based on specific engineering objects or even distinct fault patterns of the same object. The challenge in current PHM lies in selecting the most suitable algorithm from a plethora of PHM algorithms according to the specific requirements of particular objects(Zou et al., 2023). LLM possesses exceptional data processing and pattern recognition abilities. By fine-tuning pre-trained models using PHM algorithms and case datasets along with relevant expert knowledge, key features within input data can be extracted and learned. This process establishes a correlation model between input data and output decisions. Leveraging the model's outstanding pattern recognition capabilities enables informed decision-making. This approach presenting an opportunity for targeted PHM algorithm intelligent recommendations in specific environments, reducing time and resource costs associated with the PHM design process.

The intelligent recommendation of PHM algorithms based on the parallelism of LLM and PHM model first necessitates the construction of a PHM algorithm case repository for PHM model. This repository incorporates a variety of algorithm configurations and usage conditions employed in actual engineering applications within the PHM field. Utilizing the knowledge extraction and internalization capabilities of LLM, algorithm configuration elements, diagnostic knowledge elements, and engineering condition elements described in these cases are transformed into PHM engineering experience through techniques such as SFT and RLHF, forming a foundation of PHM algorithm knowledge. Subsequently, the accumulated PHM experience within the LLM is harnessed to intelligently recommend algorithms based on user-interacted prompts that outline PHM task requirements(L. Li et al., 2023; J. Liu, Liu, et al., 2023; Nikzad-Khasmakhi et al., 2019; L. Wu, Zheng, et al., 2023). During user engagement, several dimensions of requirements must be determined, including the type of usage object, the specific PHM task dimension (e.g., fault diagnosis, life prediction, health assessment, maintenance decision), available data and computational resources, and anticipated outcomes. Through interactive dialogues with the LLM, the most suitable algorithm & model and relevant configuration elements for the current PHM requirements are recommended.

In order to elevate the recommended algorithms and related configuration elements from a stage of *credible and usable* to a leading stage of *highly reliable and user-friendly*, continuous evolution is essential within the paradigm of parallel development involving LLM and PHM model. Specifically, LLM could potentially learn inappropriate or even incorrect PHM engineering experience due to issues with the quality of algorithm cases, leading to serious misconceptions(Lee et al., 2021)(Ji et al., 2022). Addressing this concern, aligning the LLM recommendations with human expert diagnostic preferences becomes a key factor in enhancing the LLM's capabilities in PHM algorithm recommendation within this specialized field. The parallel development of PHM model allows for the simultaneous validation of the efficacy of PHM algorithms recommended by the LLM. Based on the algorithms & models and related configuration elements recommended by the LLM, the PHM model is constructed. By evaluating the practical performance of algorithms, such as diagnostic accuracy, prediction precision, and fault isolation rate, through actual operations, feedback is obtained to assess the quality of the PHM algorithms recommended by the LLM. Recommended PHM algorithms by the LLM should yield optimal performance metrics after real-world implementation; if not, these instances of error are used to deeply align the LLM to rectify inaccuracies for the current application. Through ongoing interaction with users, the precise PHM algorithm recommendation capability of the LLM continues to improve. In summary, the PHM algorithms and configuration elements recommended by the LLM can be positively advanced through continual interaction, feedback, and iteration with humans and PHM model.

### 4.3.3 Approach 3: Logical Support of PHM Algorithm & model

**Name:** **Logical Support of PHM Algorithm Model**

**Problems:** **Algorithm model lacks logical inference, poor credibility**

**Advantages of LM:** **Problem analysis, knowledge inference, text generation**

**Inputs:** **PHM algorithm analysis process, object knowledge characteristics**

**Outputs:** **Algorithmic model reasoning process, interpretation of results**

**Typical application forms: logical supplementation of black-box algorithms, development of highly trustworthy PHM algorithms**

**Fig. 4-10 Information Card of Logical Support of PHM Algorithm & model**

The approach of logical support of PHM algorithm & model aims to address the issues of lack of logical reasoning and poor credibility of black-box algorithms & models. Guided by the idea of Paradigm II that LLM lead PHM model to achieve parallel development, this approach leverages the strengths of LLM in problem analysis, knowledge reasoning, and text generation. It achieves the task of providing logical support for PHM algorithms & models by taking the process of PHM algorithm analysis and the characteristics of the objects' knowledge as inputs, and producing the reasoning process and results interpretation of the models. This approach is geared towards improving the logical coherence and credibility of algorithms & models in the PHM field, through applications like supplementing the logical coherence of black-box algorithms & models and developing highly credible PHM algorithms.

Logical coherence refers to the rationality and rigor of an argument, reasoning, or decision-making process. It is one of the essential criteria for assessing the credibility and acceptability of an argument or viewpoint. In machine learning, many models, such as neural networks, are considered black-box models. The algorithms within these models operate internally in complex and opaque ways. We are unable to directly observe or understand the decision-making logic of the algorithms, making the results of the model difficult to interpret and comprehend. It also becomes challenging to verify whether the algorithm makes decisions based on reliable data and correct assumptions, thus raising doubts about the credibility of black-box model results(Durán & Jongsma, 2021). However, data-driven algorithms analyze and process a large amount of data to extract useful information for visualization, enhancing users' understanding of the algorithm's process. This can increase the credibility of black-box models. LLM is supported by extensive text data, which possesses strong knowledge reasoning capabilities, enabling it to deeply learn and reason about knowledge. During the reasoning process, LLM analyzes and reasons through problems logically, generating more comprehensive and in-depth answers, thereby enhancing the logical process of results. Relying on the data processing capabilities of data-driven algorithms and the knowledge reasoning capabilities of LLM, it's possible to optimize the design and selection of black-box models. This can improve the logical coherence and address the limitations of result credibility, ultimately assisting PHM algorithms in achieving highly credible diagnosis, prediction, assessment, recommendation, and decision-making processes.

Combining the data processing capabilities of data-driven algorithms with the knowledge reasoning capabilities of LLM significantly strengthens the logical coherence support of PHM algorithms & models. Data-driven algorithms & models analyze and process a large amount of data, employing operations such as feature extraction, feature engineering, and visualization to process the data. These models can also utilize interpretable methods like decision trees to help us better understand the decision-making process and outcomes of the models. During knowledge reasoning, LLM parses and comprehends the input textual information, performs logical analysis and reasoning during the reasoning process, and establishes reasoning chains using relevant knowledge points. For instance, in the context of a fault-tree model, the model could reason from the bottom-level units to the top-level fault conditions by considering the series and parallel relationships between different levels. LLM can also engage in analogical reasoning, comparing against existing knowledge repositories to infer solutions to current problems, thereby enhancing algorithm generalization and logical coherence. Furthermore, developing appropriate evaluation metrics and generating explanatory outputs allows LLM to provide reasoning along with their outputs, enhancing their logical coherence at the knowledge level. By integrating the capabilities of mature LLM with PHM model, and by fusing knowledge and data, the logical coherence and credibility of PHM algorithms & models can be further enhanced. This comprehensive approach ensures that the results produced by PHM model are process-oriented and interpretable.

### 4.3.4 Approach 4：PHM Algorithm & model Intelligent Optimization and Updating

**Name:** PHM Algorithm Model Intelligent Optimization and Updating
**Problems:** Algorithm model accuracy, performance degradation
**Advantages of LM:** Logical reasoning, decision-making judgment
**Inputs:** Parameters, deviations, data, conditions, feedbacks
**Outputs:** Updated algorithmic model, evaluation results

**Typical application forms: PHM algorithm development for new research products, adaptive algorithm update**

**Fig. 4-11 Information Card of PHM Algorithm & model Intelligent Optimization and Updating**

The approach of PHM algorithm & model intelligent optimization and updating addresses the issue of reduced accuracy and performance of algorithms & models due to changes in the object, operating conditions, and data conditions. Guided by the idea of Paradigm II that LLM lead PHM model to achieve parallel development, this approach leverages logical reasoning, decision-making, and robust data processing capabilities of LLM. This approach takes various elements into consideration, including algorithm & model parameters, deviations in running results, accumulated data, condition changes, and human-computer interaction feedback, among others, as inputs. The outputs of this process are intelligently optimized and updated algorithm & model and evaluation results. This approach is applied in scenarios like the developing new PHM algorithms for products, as well as adaptive updating of algorithms during operational service periods. By integrating these techniques, the approach contributes to a comprehensive enhancement of the versatility and adaptability of PHM algorithms & models.

In the field of PHM, the configuration of algorithm & model parameters plays a crucial role in determining algorithm performance. Current PHM algorithms & models often cater to specific objects or operating conditions, and their performance is unstable and generalization is limited in the face of different objects(Y. Hu et al., 2022), varying operating conditions, or diverse data conditions. In practical engineering applications, as the application objects of PHM algorithms & models change, the corresponding operating and data conditions may also change. Consequently, the performance of the original algorithm & model might deteriorate in such scenarios, necessitating updating and optimization to ensure its effectiveness. Moreover, system generates a significant amount of new data during usage, demanding the PHM algorithms & models to be updated and optimized based on such new data to maintain their performance and accuracy(Z. Zhang et al., 2018). Current PHM algorithms & models often require manual parameters adjustment or heuristic algorithms to optimize parameters when applied to new environments or datasets. This manual adjustment process relies heavily on experiential knowledge, which may introduce human errors through trial-and-error parameter configuration attempts. Using heuristic algorithms to optimize parameters demands customized design of model constraints and objective functions, which can be labor-intensive. Additionally, the data in the PHM field typically exhibits high dimensionality and complexity, resulting in a large parameter space for algorithms(W. Zhang et al., 2019). This makes parameters tuning even more challenging and limits the generalization of PHM algorithms. LLM possesses strong generalization capabilities. When combined with techniques like fine-tuning and RLHF, it can achieve intelligent optimization and updating of model parameters based on different new data conditions.

The intelligent optimization and updating of PHM algorithms & models based on the parallel development of LLM and PHM model follow a specific process. Firstly, the PHM model calculates the current algorithm & model's accuracy and deviation based on the computed results and actual diagnostic, predictive, evaluative outcomes. When the accuracy deviation of the algorithm & model is significant and cannot meet the requirements of PHM, the deviation value and various algorithm parameters of the algorithm & model are input into the LLM. Leveraging the knowledge extraction and reasoning capabilities of the LLM and referring to similar algorithms & models and their parameter settings in the PHM model dataset, the LLM provides optimization suggestions for the current algorithm & model's parameters. The PHM model then takes the provided suggestions to update the algorithm parameters and evaluates the updated algorithm & model. This evaluation aims to verify whether the accuracy of the new algorithm & model calculations has improved and whether it can fulfill the requirements of PHM. If the updated model meets the requirements, the optimization and updating of the algorithm & model for that iteration is complete. However, if the model still fails to meet the requirements, feedback is provided to the LLM regarding the current optimization and updating status. New parameters optimization suggestions are generated, and the process of algorithm & model optimization and updating will be repeated iteratively until the PHM requirements are fulfilled. This approach ensures that the PHM algorithms & models continuously improve and adapt to changing conditions and data.

## 4.4 Paradigm III: Construction and Application PHM-LM

For requirements such as PHM data generation, PHM capability generation, complex system PHM solutions generation, and PHM verification & validation, we develop a comprehensive design of PHM-LM. This design considers the unique characteristics of the PHM field and references classic Large Model's framework, as illustrated in Fig. 4-12. Also, this design involves a step-by-step approach in the fusion of PHM and Large Model, encompassing the aspects of object dimension, task dimension, algorithm dimension, and modality dimension. Generally speaking, the design process of the PHM-LM includes tasks such as model development, data collection and annotation, pre-training, fine-tuning, model alignment, model evaluation, and inference. This research work aims to enhance PHM capabilities, address challenges within the PHM field, optimize Paradigm I and II, and ultimately construct the following types of Large Model:

1. Object-oriented PHM-LM, such as those for bearing, gear, motor, electronic product, and more.

2. Task-oriented PHM-LM, such as those for data generation, algorithm recommendation, solution generation, verification & validation, and more.

3. Algorithm & model-oriented PHM-LM, such as those for fault diagnosis, fault prediction, maintenance decision, and others.

4. Multi-modal PHM-LM, such as those for vibration modal, knowledge modal, text modal, image model, multi-modal, and more.

These types of Large Model are designed to cater to a range of specific requirements in the field of PHM, ensuring a comprehensive approach to enhancing capabilities, solving bottlenecks, and optimizing various aspects of the PHM field.

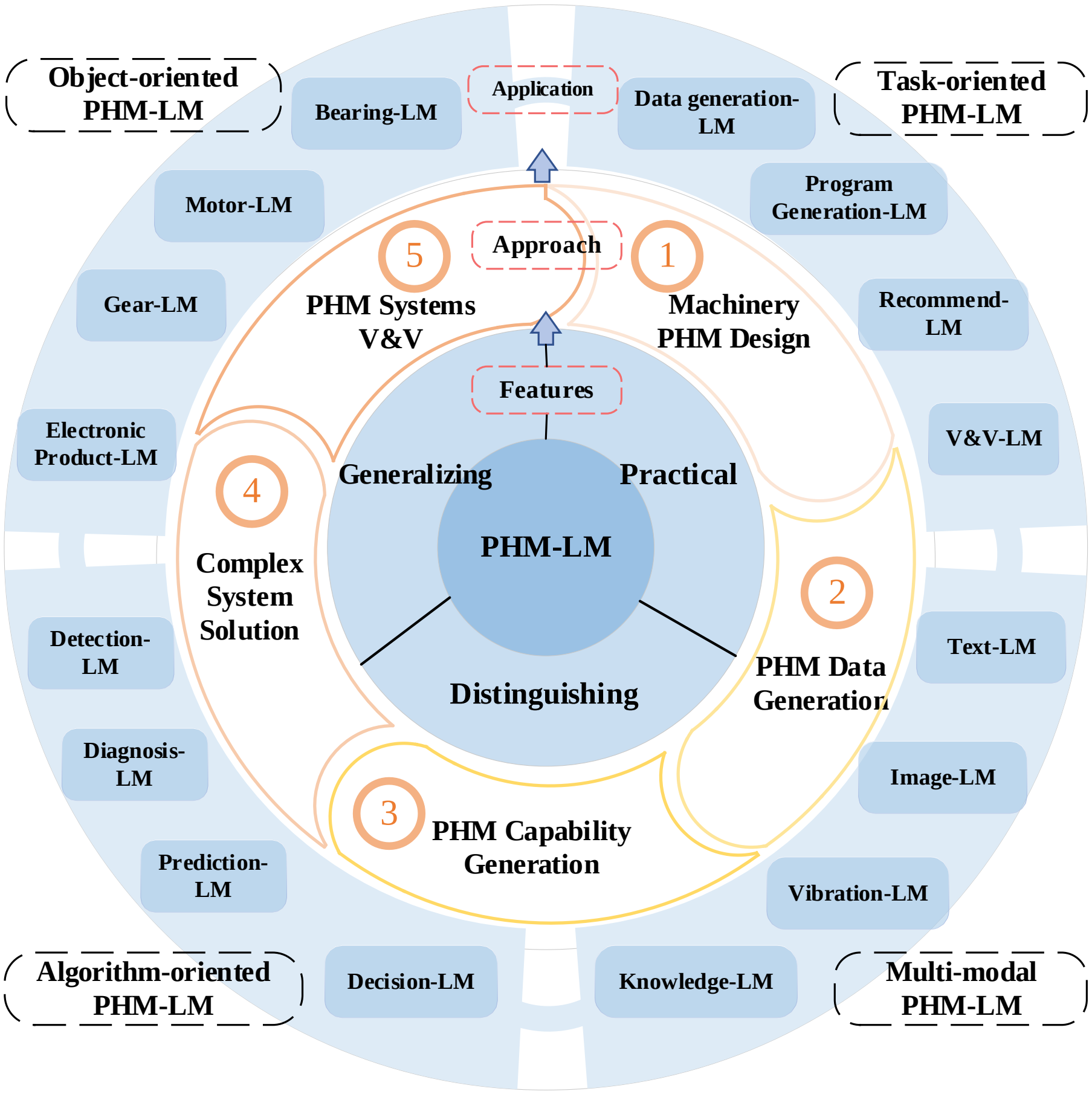

**Fig. 4-12 Construction and Application PHM-LM**

### 4.4.1 Construction of PHM-LM

Traditional PHM techniques often only form capabilities specific to a particular object, but, in general, they still struggle to scale and emerge across dimensions such as processes, functions, and objects. Currently, the PHM field faces bottleneck issues like improving generalization performance, enhancing interpretability and credibility, establishing a verification system, lowering design and development thresholds, efficiently utilizing field knowledge, and boosting interactive capabilities and efficiency. Paradigms I and II leverage the framework and functionalities of existing LLM, mitigate and solve the problems faced by PHM to some extent by addressing the specific multimodal data, differentiated tasks across stages, and entire-process functionalities in the PHM specialized field. However, a comprehensive and fundamental technological innovation is imperative for a definitive solution to these challenges.

Drawing inspiration from the text generation capabilities of LLM that use natural language as their elemental component, if one could harness the emergent and generalizing features of Large Model, we could construct a PHM-LM from the ground up. This model would be based on multi-modal information from the PHM field, centered around various PHM functional needs, and aimed at generating emergent capabilities. Such a model would bridge the technical processes of PHM development solution generation, data generation, capability generation, solution generation, verification & validation. In addition, this supports downstream tasks such as PHM design, diagnosis, assessment, prediction, decision-making, recommendation, verification. This approach would not only address challenges faced by traditional PHM but also realize a fundamental technological shift in the PHM field characterized by generalization, discriminative abilities, and practicality.

From a construction perspective, building a Large Model involves steps like designing the Large Model architecture, large-scale pre-training, fine-tuning instructions for different objects/tasks/conditions, and enhancing inferencing capabilities through reinforced guidance. As per the law of expansion(Kaplan et al., 2020),, enlarging the model or dataset typically improves the performance of downstream tasks, making data volume foundational to the emergent capabilities of Large Model. The information characteristic in the PHM field is multi-modal, and the complementarity between multiple modalities can enhance model perceptual abilities(Yin et al., 2023). Therefore, building a generic generative PHM-LM necessitates preparing extensive, multi-modal PHM field data for the model's pre-training and fine-tuning phases. To ensure data quality, appropriate preprocessing strategies(Rae et al., 2021) should be employed, including quality filtering, deduplication, privacy removal, and labeling. Meanwhile, significant manpower is required to tokenize every modal information type. For vast PHM field data and larger model scales, most existing research resorts to batch training with dynamic adjustments. The construction of a PHM-LM entails steps like hyperparameter search, large-scale distributed training, and production inference, encompassing cluster design, rack design, and node design. Facing trillions of parameters, training Large Model requires parallel processing across multiple machines and cards. Hence, tasks, training data, and models are segmented for distributed storage and training.

Subsequently, based on the PHM-LM, and centered around classic PHM tasks such as end-to-end PHM design, solution development, PHM data production, PHM capability creation, complex system PHM solution formation, and PHM verification & validation, this section will systematically elaborate on the approaches and potential technical paths that the PHM-LM may adopt to resolve various challenges.

### 4.4.2 Approach 1：Intelligent End-to-End PHM Design

**Name:** **Intelligent End-to-End PHM Design**
**Problems:** **Consuming a lot of manpower and relying strongly on experts' experience**
**Advantages of LM:** **Reasoning ability, multi-modal fusion ability**
**Inputs:** **Task information, requirement information, historical cases**
**Outputs:** **Full-process PHM design program**

**Typical application forms: automatic design and generation of full-process PHM solutions**

**Fig. 4-13 Information Card of Intelligent End-to-End PHM Design**

The Intelligent End-to-End PHM design route seeks to address the significant human resource consumption and heavy reliance on expert-driven design in traditional PHM design. Within the PHM-LM architecture constructed in Paradigm III, the design capitalizes on the Large Model's inferential capabilities and large modal fusion abilities. It facilitates an intelligent PHM design process where task details and historical PHM design cases are inputs, and the result is a comprehensive PHM design. This approach promotes the rapid intelligent evolution of PHM design through applications such as end-to-end PHM development and design.

One of the core functionalities in the PHM field is custom design. This encompasses fault diagnosis, health assessment, and life expectancy prediction across various stages(Baur et al., 2020). Typically, an end-to-end PHM design mandates that PHM requirements are analyzed during the requirement analysis phase. During the conceptual design phase, the PHM solution undergoes verification & validation. In the preliminary or detailed design phase, the corresponding design is formulated, with functional models clearly defined. The detailed design and assessment are finalized during the detailed design phase, and system evaluations and design typing are conducted in the prototype testing phase. Throughout the mass production and service phase, functionality is validated, applied, and feedback for design improvements is gathered. Historically, the PHM design for different types of systems in varying application scenarios has been predominantly driven by expert experience and those system characteristics, leading to designs with limited generalizability(Z. Liu et al., 2018). To propel the evolution of PHM design towards generative, intelligent, general, and end-to-end directions, there's a pressing need to intelligently generate entire-process PHM solutions based on the PHM specialized field Large Model. This would alleviate the heavy reliance on expert knowledge in PHM system design and truly advance the transformation of intelligent technologies in the PHM field.

In the design of the end-to-end PHM solution based on the Large Model, the primary dependency is on requirement analysis. The Large Model, when provided with detailed task information (including operational support requirements, usage scenarios, high-cost protection activities, and other key activities), should leverage its inferential capabilities to perform requirement analysis. This produces overarching technical requirements and preliminary validation, aligning them with relevant PHM system norms and system validation reports. Expectedly the PHM-LM, during target object PHM design, should intelligently analyze metrics distribution requirements, system functional needs, and service maintenance requirements at every phase. This would automate the creation of status monitoring, data processing, algorithms & models, integrated coupling, validation and evaluation, and other specific solutions for technical personnel reference and application. In this context, the Large Model, through vectorization and emergent capabilities, can present intelligent matching operations between and within phases in an inferential manner, conserving intellectual and manual efforts from technical staff. A pivotal aspect of the end-to-end design is addressing strong constraint conditions. In the PHM design process, besides the PHM functionality, there are interactions and constraints involving the system's primary design, external datasets, and data resources. This forms the constraining input for the entire design process under the overarching PHM requirements and capabilities. Given the extensive timeline of the end-to-end PHM solution, enhancing the alignment of model inferences with actual scenarios necessitates setting up evaluation interfaces at multiple nodes throughout the PHM process. This ensures the effectiveness and reliability of the PHM development solutions produced by the Large Model during practical application. Consequently, end-to-end PHM design requires multi-phase training and fine-tuning, adopting a modular design approach in tandem with the PHM design stages. Additionally, through in-built evaluation algorithms and external human cues, the model's generative capabilities are optimized.

### 4.4.3 Approach 2：Intelligent PHM Data Generation

**Name: Intelligent PHM Data Generation**

**Problems: Algorithmic modeling needs of high-quality data is difficult to meet**

**Advantages of LM: Strong generation capability, strong generalization performance**

**Inputs: Information on object equipment, knowledge of data characteristics**

**Outputs: Simulation data under predetermined working conditions and states**

**Typical application forms: typical failure mode data generation, validation data profile construction**

**Fig. 4-14 Information Card of Intelligent PHM Data Generation**

The intelligent PHM data generation approach aims to address the challenge of meeting the high-quality data requirements for PHM algorithms & models. Within the Paradigm III management Large Model architecture, the generation capabilities and generalization performance of Large Model, it realizes the task of generating simulated data that can mimic the predetermined state of system under predetermined operating conditions and environments. By utilizing typical applications such as generating fault data and constructing machinery task validation data profiles, it provides convenience for the design, development, and validation of PHM algorithms & models, promoting a comprehensive enhancement of PHM algorithm & model capabilities.

PHM data plays a crucial role as input for PHM design, development, and algorithms & models within the PHM technical system. Having good data conditions can help achieve accurate, fast, and reliable PHM capabilities. However, in practical applications, PHM data is often affected by complex operating conditions and harsh usage environments, leading to issues such as data missing, data imbalance, and insufficient information content, which severely impact the effectiveness of PHM development, testing, validation, and utilization. Currently, data augmentation techniques serve as a core solution to address the challenges of insufficient data for data-driven algorithms & models in the field of PHM and even artificial intelligence. Although generative network models, data transfer, and variational encodings have alleviated the conflict between the demand for large amounts of data for algorithms & models and the suboptimal conditions of real-world engineering application data to some extent, the generated data still lacks consideration for factors such as objects, operating conditions, and future application requirements, resulting in significant differences from real-world data. To address this issue, the PHM-LM will comprehensively optimize the data generation task from the perspectives of input considerations, analysis processes, and output forms, following the specific workflow outlined below:

In the framework of the PHM-LM, a generative pre-training model is trained using historical data from various objects under their respective operating conditions and environments. The model considers core elements such as design information, application environment information, and future algorithm & model characteristics at the input stage, establishing associations between the generative model and these core elements through large-scale pre-training. Furthermore, considering the characteristics of the data, the model enhancement process is guided based on levels of consistency, diversity, and task-specificity, aiming to improve the similarity between the generated data and real data. This ensures the correctness and credibility of the data generation process. Finally, by incorporating the PHM design elements within the Large Model framework, the PHM data generation output is constructed based on the PHM design algorithm development requirements. Validation metrics and algorithm cases are designed to verify the generated data and provide a quality assessment of the PHM generated data.

The application of the generative capabilities of the PHM-LM in PHM data generation tasks can provide reliable data generation considering diverse modalities such as machinery characteristics, operating conditions and environmental differences, design elements, and application requirements. The generated data will outperform existing data generation techniques in terms of consistency with real data, diversity in generated data, and adaptability to specific tasks. The comprehensive improvement in both the quantity and quality of PHM data facilitates the design and development of PHM algorithms & models, and provides avenues for the verification of PHM algorithms and the evaluation of PHM systems at various levels, which were previously challenging due to the lack of sufficient data support.

### 4.4.4 Approach 3：Intelligent PHM Capability Generation

**Name: Intelligent PHM Capability Generation**

**Problems: High threshold of traditional development and low degree of intelligence**

**Advantages of LM: Strong generation capability, strong reasoning capability**

**Inputs: Object equipment information, capability requirement elements**

**Outputs: Formation of PHM capability that meets development requirements**

**Typical application forms: intelligent generation of diagnostic capability, intelligent generation of predictive capability, etc.**

**Fig. 4-15 Information Card of Intelligent PHM Capability Generation**

The intelligent PHM capability generation approach aims to address the challenges of high barriers and low level of automation in the PHM capability development process. Within the Paradigm III management Large Model architecture, leveraging the generative capabilities and inference performance of the Large Model, it aims to develop PHM capabilities that meet the development requirements by taking system information and capability demand factors as input. Typical applications such as diagnostic capability generation and prognostic capability generation significantly reduce the technical barriers associated with PHM capability development and enhance the level of automation in the capability generation process. This approach promotes a comprehensive transformation in PHM capability generation technology.

The implementation of PHM capabilities relies on the underlying development of PHM algorithms & models, and excellent underlying code is key to ensuring the effectiveness of PHM capabilities. However, in practical PHM development processes, developers often struggle to independently carry out the tasks of developing underlying code. Most development tasks require task descriptions to be provided to professional programmers who organize the development process. Due to the relatively limited field knowledge in the field of PHM among professional programmers, the exchange of information between developers can be inefficient. Additionally, the developed code may deviate from the anticipated requirements and have low accuracy in the algorithms & models. Currently, advanced PHM design support systems in China have implemented the functionality of secondary development based on existing encapsulated code. Users can secondary develop the underlying code through a modular approach, which has to some extent reduced the threshold for PHM design and development. However, the task of secondary development still requires configuring the interaction between modules and a thorough understanding of the functionality of each code module. Otherwise, design errors are likely to occur during the development process, making debugging difficult. In the context of Large Model technology, the generative capability of Large Model provides a new approach for intelligent PHM capability generation

Based on the PHM-LM, intelligent PHM capability generation mainly consists of two functions: intelligent generation of bottom-level code and intelligent generation of parameter configuration information for PHM algorithms & models. The intelligent generation of bottom-level code utilizes historical PHM algorithm & model codes as training foundations, forming task-specific code frameworks for core tasks such as diagnosis, prediction, assessment, validation, and recommendation. During the generation of bottom-level code, the overall development requirements and task characteristics of the system's PHM capability are taken into consideration, considering factors such as object characteristics, operational conditions, environment, and data format. By leveraging the reasoning and generation capabilities of the Large Model, the bottom-level code that meets the task requirements can be quickly generated. Subsequently, through the analysis of algorithm & model characteristics and data information, the optimal parameter configuration for the PHM algorithm & model is determined to ensure the task accuracy of the intelligent-generated PHM bottom-level code, enabling efficient and precise generation of PHM capabilities.

With the increasing demand for PHM system usage and the growing complexity of PHM algorithms & models aiming for better performance, the generation of PHM capabilities has become increasingly challenging. Under the influence of the generative capabilities of Large Model in PHM, combined with the reasoning abilities of these models, it becomes feasible to achieve fast and accurate generation of algorithm & model bottom-level code and capabilities in an automated and standardized manner. This approach is essential in reducing the threshold for PHM development, shortening the time required for PHM capability generation, and minimizing the resources and costs associated with PHM algorithm development.

### 4.4.5 Approach 4: PHM Solution Generation for Equipment Systems

**Name:** **PHM Solution Generation for Equipment Systems**
**Problems:** **Low multi-modal information processing capability**
**Advantages of LM:** **Multi-modal information processing, logical reasoning**
**Inputs:** **Parametric indicators, structural information**
**Outputs:** **PHM solution**

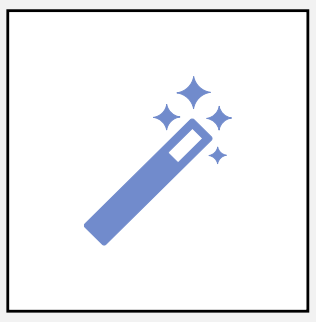

**Typical application forms: PHM program generation for equipment systems**

**Fig. 4-16 Information Card of PHM Solution Generation for Equipment Systems**

The route for generating equipment system PHM solutions aims to enhance the multi-modal information processing capability during the PHM design phase. Guided by the Paradigm Ⅲ of the PHM-LM model, this approach leverages the advantages of multi-modal information processing, logical reasoning, and natural language generation. It takes multiple factors such as multi-modal fault information, structural information, object characteristics, application scenarios, operational conditions, parameter indicators, and capability requirements as inputs to generate PHM solutions. The generation of PHM solutions for equipment systems serves as a typical application that enhances the ability of PHM technology to address the challenges of interdependent and interconnected equipment systems.

Traditional PHM solution designs often focus on a single type of information and neglect the processing of multi-modal information. By only considering specific sensor outputs and neglecting other modalities such as sound, vibration, and temperature, the solution design may be incomplete. In addition to multi-modal input information, traditional PHM designs often overlook the importance of structural information. Equipment systems are typically composed of multiple components, and these components exhibit intricate interconnections and couplings. This means that any changes in one component can potentially impact others, or even the entire system. Therefore, for PHM solutions targeting equipment systems, it is crucial to consider these interconnections, couplings, and their impacts on the overall system health. Object characteristics are also often overlooked, as different systems, components, or software may exhibit their unique operating characteristics and failure modes. If the PHM solution fails to account for the specifics of each object, the accuracy of prediction and diagnosis will be greatly compromised. Additionally, the importance of application scenarios, operational conditions, and parameter indicators in PHM design cannot be underestimated. These indicators provide quantifiable information about the system's health state. In summary, to effectively address the various challenges in equipment systems, a comprehensive and integrated PHM solution is needed. This solution should not only focus on the operation of individual components but also leverage multi-modal information, structural information, object characteristics, application scenarios, operational conditions, and parameter indicators to ensure the health and stable operation of the entire system.

In this context, the adoption of PHM-LM model undoubtedly provides strong assurance for the efficient and stable operation of systems. The introduction of PHM-LM models can effectively leverage the information from the object and algorithms & models to generate a comprehensive PHM solution. Specifically, Large Models possess the capability of processing multi-modal information, including text, sound, images regarding systems, as well as structured sensor data. This provides a solid foundation for data-driven analysis in PHM solutions. Additionally, Large Model exhibit powerful logical reasoning capabilities. In the operation of equipment systems, small variations can trigger a series of cascading effects. Through logical reasoning, Large Model can not only track the relationships among these variations but also integrate them with historical and similar object information. On the other hand, the text generation capability of PHM-LM models can convert complex analytical results into clear and concise textual descriptions. Whether it's the current operating status and parameter information of the system or the information of the algorithms & models, they can be organized into explicit and coherent PHM solutions.

### 4.4.6 Approach 5: Intelligent PHM Verification & Validation

**Name:** **Intelligent PHM Validation & Evaluation**

**Problems:** **Lack of validation evaluation of traditional PHMs**

**Advantages of LM:** **Logical reasoning ability**

**Inputs:** **Structural information, failure modes**

**Outputs:** **Validation results and evaluation**

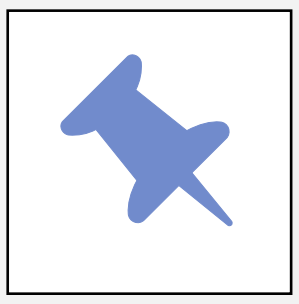

**Typical application forms: PHM verification & validation**

**Fig. 4-17 Information Card of Intelligent PHM Verification & Validation**

In order to address the lack of verification & validation in traditional PHM, the intelligent PHM verification & validation route, guided by the Paradigm Ⅲ of PHM-LM models, leverages the logical reasoning capabilities of Large Model. It takes system structural information, input information, and fault patterns as inputs to generate verification results and evaluations. This approach brings new technological breakthroughs to the field of PHM through typical applications of PHM verification & validation.

In traditional PHM Verification & Validation, it is challenging to establish unified evaluation criteria due to the complexity and diversity of system. Different devices may have different operating principles, system parameters, application environments, and fault patterns, making it difficult to develop a general evaluation standard for them. Moreover, traditional PHM methods often rely on the expertise of technical personnel. While this expertise can be valuable in certain cases, it can also introduce biases and misconceptions. On the other hand, extensive fault data is required to validate the accuracy of PHM methods. However, in many situations, this data is either unavailable or difficult to obtain. For instance, some devices may experience rare failures, or the fault data may be considered sensitive and not publicly released. This makes thorough validation of PHM methods challenging. In PHM evaluation, effective evaluation metrics are necessary to assess the performance of PHM methods. However, these metrics may not be applicable to all devices and applications. Therefore, it is essential to develop more suitable evaluation metrics tailored to different devices and applications.

PHM-LM models have the ability to understand and process a vast amount of information. Compared to traditional methods, Large Model possess stronger logical reasoning capabilities, enabling them to tackle more complex problems and provide more accurate and objective verification & validation. Leveraging the advantages of Large Model, we can apply them to the evaluation of PHM. By inputting system structural information, input information, and fault patterns, the PHM-LM model can quickly grasp the working principles of the devices and potential causes of failures. By simulating the device's operation, it can make predictions and diagnoses and compare them with actual data to validate the accuracy of PHM results. Additionally, the Large Model can provide evaluations for PHM methods. By analyzing the actual operating data of the system, the model can assess the performance of PHM methods based on evaluation metrics. These metrics can help us understand the strengths and weaknesses of the PHM methods, facilitating further optimization.

# 5. Application Deployment Mode for PHM-LM

This chapter delves deeply into the deployment strategy of PHM-LM models, which are deployed in cloud servers. The strategy is mainly divided into three levels: the foundational layer, industry-generic layer, and specialized application layer. As shown in the Fig. 5-1, the foundational layer serves as the pillar of the entire deployment paradigm, providing a robust architecture. This architecture not only establishes the basic structure of the PHM-LM model but also defines its application characteristics, ensuring the smooth implementation and application of the subsequent two layers. The industry-generic layer is endowed with more specific and extensive functionalities. For instance, multi-modal data processing techniques ensure that data collected from various sources is effectively utilized; algorithm generation technology allows the PHM-LM model to autonomously generate and optimize its functionalities. Meanwhile, the generation, verification & validation ensure that the proposed PHM solutions are always in their optimal state. This layer furnishes the PHM-LM model with a comprehensive set of tools and methods, enabling it to better serve the specialized application layer.

The specialized application layer is exceptionally critical, as it directly interfaces with the model's end-users and developers. Contrasting with the theoretical and technical nature of the preceding two layers, this layer places a greater emphasis on practical applications and user experience. From aviation to high-speed rail, and from wind power to automobiles, every application field possesses its unique requirements and characteristics. To cater to these needs, the specialized application layer offers developers a series of API interfaces, allowing them to adjust and optimize the model according to real-world situations. Additionally, to ensure that general users can also effortlessly use it, this layer specifically designs a concise and clear user interaction interface.

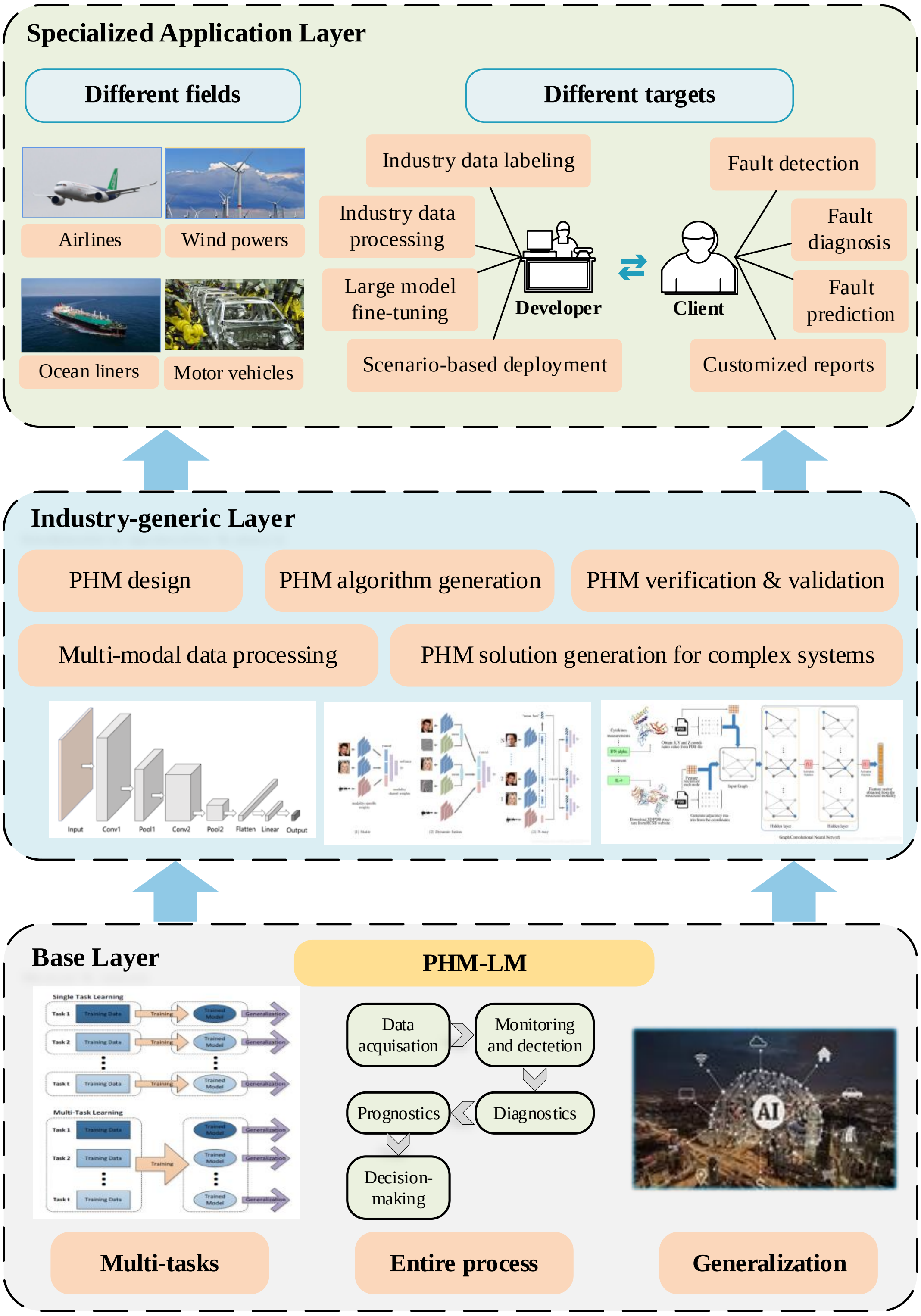

**Fig. 5-1 PHM-LM Application Deployment Model**

## 5.1 The Foundational Layer

The foundational layer serves as the core pillar for the PHM field, characterized by multi-task learning, entire-process support, and general adaptability. It facilitates multi-task sharing, covers the entire process from data collection to decision-making, and possesses generalization to fine-tune both the general layer and specialized application layer. The foundational layer furnishes the industry-generic layer and the specialized application layer with robust algorithm and computational power support, ensuring the stable operation and long-term reliability of the PHM-LM model.

Multi-tasks: PHM doesn't solely concern the health status assessment of a system. It encompasses various tasks, including fault detection, fault type classification, and the prediction of remaining useful life. The foundational layer can not only support multiple tasks simultaneously but also handle and possesses multi-task learning capability. This ensures knowledge sharing between different tasks, subsequently enhancing the performance of each task.

Entire-process: The implementation of PHM isn't just about prediction and maintenance decisions. Ranging from data collection, preprocessing, feature extraction, model training, validation, testing, to fault detection, diagnosis, prediction, and culminating in PHM decision support – it covers a comprehensive process. The foundational layer must offer entire-process support to ensure seamless data and information flow, thereby delivering accurate and timely services for upper-layer applications.

Generality: Although different devices and systems may have their specific characteristics and requirements, many fundamental tasks and challenges in PHM are common. The foundational layer, trained and tested on public datasets, boasts significant generality. This allows it to be applied to different devices and scenarios in the upper layers without requiring massive customization. At the same time, the foundational layer offers an array of tools and interfaces, facilitating necessary customization and extension for upper-layer applications.

## 5.2 The Industry-generic Layer

The industry-generic layer provides comprehensive support for the application of the PHM-LM model through multi-modal data processing, algorithm generation, complex system PHM solution generation, PHM verification & validation.

Multi-modal data processing: Playing a pivotal role in the industry-generic layer, multi-modal data processing manages information from varied sensors and data sources, such as vibration, temperature, and sound. It transforms this heterogeneous data into a consistent format, ensuring comparability and coherence. Furthermore, it carries out data cleansing, feature extraction, and data fusion, thus offering high-quality input for subsequent analysis. This ability is especially crucial in managing the health of intricate devices and systems, laying a robust foundation for the model's accuracy.

Algorithm & model generation: The industry-generic layer boasts superior algorithm generation capabilities, automatically formulating algorithms suitable for analyzing and predicting from multi-modal data. These algorithms are flexibly chosen and adjusted based on distinct tasks and data features. This capability allows users to effortlessly apply advanced analysis techniques without delving into the nuances of the algorithms.

Complex system PHM solution & generation: The layer possesses the aptitude to generate predictive PHM solution for intricate systems. Leveraging the outcomes of multi-modal data processing and the auto-generated algorithms, the solution generation autonomously designs a complete PHM solution in accordance with user demands, device traits, and environmental context. This encompasses data collection strategies, monitoring blueprints, and model configurations, offering explicit operational guidance for real-world applications and diminishing user workload in design.

PHM verification & validation: The verification & validation of the PHM solution, conducted through simulation experiments or live tests, assess the solution's stability in varied scenarios. It can autonomously construct an evaluation environment, simulate different faults and conditions, thereby objectively gauging the performance of the PHM solution. This automated validation procedure vouches for the reliability and practicality of the solution, bestowing a scientific foundation for its real-world applications.

## 5.3 The Specialized Application Layer

In the PHM-LM deployment model, the specialized application layer plays a crucial role. It focuses on providing customized services to various industries such as aviation, aerospace, marine, energy, and rail, as well as developers and users of different technical levels and service backgrounds. By offering diverse interfaces and user interaction interfaces, the specialized application layer aims to achieve the following core goals:

User-oriented: Centered around users, the specialized application layer offers a simplified user interface and one-click operations for various industry fields. This makes functions such as fault detection, diagnosis, and prediction more convenient and user-friendly. With just a few simple steps, users can complete complex fault detection, diagnosis, and prediction tasks. This significantly lowers the operational barrier, allowing individuals without a deep technical background to easily use the system.

Customized reports: Users can receive tailored health reports for specific objects or systems. These reports are presented with intuitive charts and clear language, helping users better understand the status and potential issues of their objects. Leveraging the predictive capability of the PHM-LM model, it aids users in foreseeing potential malfunctions and provides maintenance recommendations. This helps in reducing system downtime and enhancing increase productivity.

Developer-oriented: The specialized application layer provides developers with user-friendly SDK toolkits and API interfaces. Through these, developers can seamlessly interact with the PHM-LM model, adapting to specific industry demands, customizing developments, and optimizing the model process. Developers can fine-tune the Large Model within the specialized application layer to meet specific field requirements. This personalized optimization allows the model to reflect real situations more accurately, enhancing PHM's precision and effectiveness.

Industry data annotation & Processing: The general layer permits developers to annotate and process industry-specific data using various tools and methods. This ensures data quality, enhancing the accuracy of the PHM-LM model.

Scenario-based deployment: It offers scenario-based deployment solutions for different industries, assisting developers in effectively applying the PHM-LM model to specific systems. Such personalized deployment ensures the efficacy and applicability of the PHM-LM model in real-world operations.

# 6. Technical Challenges for PHM-LM

PHM-LM represents a specialized industry model within the realm of PHM. In the process of evolving its technological pathways and application deployments, it is inevitable that this model will encounter the technical challenges faced by AI Large Model. As pointed out in the research report *AI Large Model Market Study (2023)*, Large Model producers must address a plethora of challenges as they progress, spanning from technical aspects, ecosystem considerations, to AI governance. These challenges encompass the substantial computational power demands, exorbitant training costs, concerns regarding data quality, limited algorithm interpretability and potential ethical and safety issues tied to AI. These elements also stand as pivotal success factors for services.

When constructing the PHM-LM, it confronts general challenges akin to those encountered in the development of other Large Models. These commonalities are primarily reflected in efficiency, deployment, and training regulations. At the same time, it will also face unique challenges distinct from those of general Large Model and those pertinent to another industry-specific Large Model. The following figure illustrates the potential technical bottlenecks the PHM-LM may face in its future evolution, which will be elaborated upon in the subsequent sections.

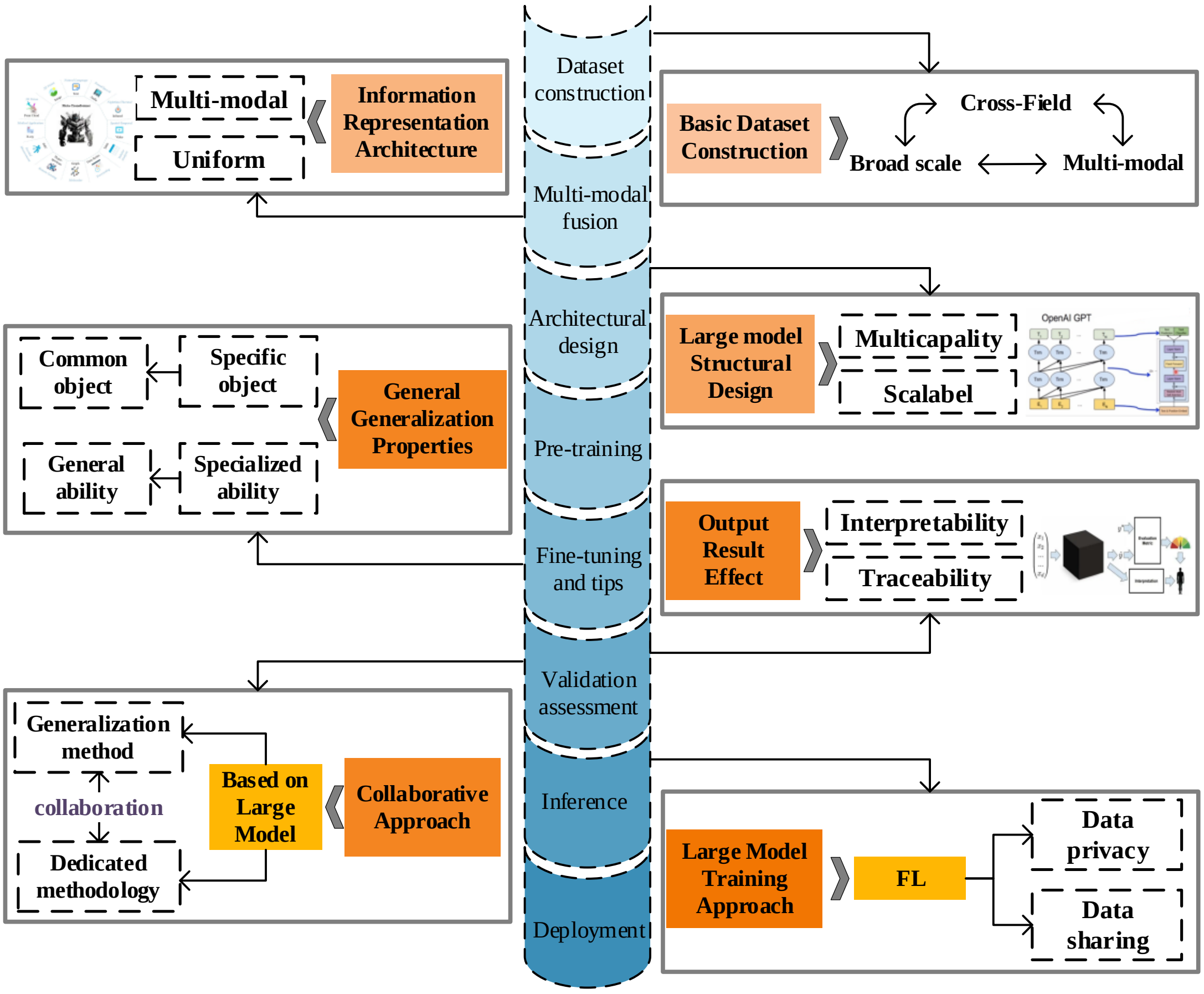

**Fig. 6-1 Sorting Out the Technical Challenges of PHM-LM**

## 6.1 Cross-field, Multi-modal Construction of PHM-LM Base Datasets

A robust dataset is fundamental for the construction of any Large Model. The PHM-LM, in particular, is characterized by its cross-disciplinary nature, multi-modality, and vast scale. On one hand, when employing traditional methods to undertake PHM tasks, there's an overlap with various technical fields beyond just PHM expertise, such as the physical mechanisms of the subject, performance metric allocation, mathematical signal processing, and system control design. Hence, data limited to just the narrow confines of PHM isn't enough to fulfill the resource requirements of a Large Model in this specialized field. On the other hand, the data in the PHM field spans multiple modalities. Specifically, this includes temperature and pressure data which change slowly, vibration data known for its rapid dynamic shifts, and voice data, commonly referred to as sensor data. In terms of knowledge, it encompasses textual information related to maintenance and repair, regulations, rules, as well as fault knowledge like

FMECA and FTA. Additionally, there's knowledge stored in other forms, like established physical laws. Case-wise, mirroring medical cases, the PHM field houses diagnostic and predictive case studies for various subjects in particular scenarios, each with its unique presentation and style.

Therefore, the collected data should cover a diverse range of application fields, scenarios, and signal modalities, and should be constantly updated and enriched to support the fine-tuning and alignment of the Large Model. Yet, based on the reference ratio concerning model size, data volume, and overall computation(Wei et al., n.d.), the data volume for a Large Model should ideally be at least at the 7B level. Given the constraints and needs of multi-modal and cross-disciplinary data, procuring such a massive volume of data for the PHM field comes with its set of challenges, like privacy concerns and specificity issues. Industry barriers might hinder access, and data regarding the specific capabilities of individual subjects could vary widely. Moreover, the combination of privacy concerns, specific subjects, functionalities, and conditions can lead to data fragmentation. In an environment where the PHM field lacks readily available datasets and APIs, finding foundational datasets becomes a significant hurdle.

## 6.2 A Unified Representation Architecture Including Multi-Modal Health Information

The prominent characteristic of health information representation in the PHM field is its multi-modality. As mentioned earlier, the multifaceted nature of PHM encompasses data, knowledge, and case studies. Current LLMs typically convert input information into vector formats for storage and use, aligning with the foundational logic of computer computations. Large Model, such as GPT-4, segment textual content or corpus based on tokens and then transform it, employing vector formats to support NLP predictions and label reward mechanisms. Such architectures not only unify data representation, simplifying model structures and enhancing computational efficiency, but also encompass a wider range of modalities, amplifying the model's generative capacity as well as its transfer and generalization capabilities. Given the multi-modal nature of health information in the PHM-LM, a unified representation architecture is both essential and beneficial. In practice, one could leverage the vector transformation strategies of existing Large Model, which would involve harmonizing the distinctive multi-modal information of the PHM field into a cohesive representation. Current research in multi-modal processing is still in its exploratory phase. For processing the second modality beyond text, Large Model generally adopt strategies such as adapting the second modality to a text modality, or re-labeling the second modality and introducing an extended dimensionality reduction vectorization layer. Whether the PHM-LM should borrow from these existing strategies or innovate a new unified representation architecture for its multi-modal information is a pressing problem that needs to be addressed during its construction.

## 6.3 Multi-capable, Scalable Structural Design of PHM-LM

When constructing a Large Model for the specialized field of PHM, this article designs multiple technical routes based on various anticipated capabilities. The realization of these capabilities should ideally be based on the foundational architecture of the PHM-LM under projected scenarios. Therefore, during the construction of the PHM-LM, there should be thorough consideration and proactive planning for multiple capability requirements like diagnosis, prediction, and assessment, as well as multiple operational demands such as real-time monitoring, maintenance, and repair. The ideal PHM-LM architecture should not only support the designed capabilities but also possess extensibility to cater to user needs beyond the initial plan, constantly evolving in tandem with technological advancements. To achieve the aforementioned objectives, the design and construction phase must integrate multiple capability generation requirements across areas such as multi-modal data representation, model utilization, training inference, command fine-tuning, and human prompts. The goal is to make the underlying architecture general, broaden the understanding of modalities, modularize inference application algorithms, and standardize input-output interfaces. Approaching from the implementation perspective of the PHM-LM, it's also essential to select an appropriate foundational architecture to reduce dependency on computational power.

## 6.4 Transformation of Exclusive Capacities for Specific Targets to General Capacities for Generalized Practice in PHM

Due to the unique nature of PHM, the construction and capability generation process of the PHM-LM will inevitably evolve from specific capabilities (such as diagnosis and prediction) for specific products like bearings and electromechanical systems, to entire-task capabilities spanning multiple-level objects. For instance, leveraging the emergent and generalization capabilities of Large Model, a specialized Large Model for diagnosing bearing faults can be developed that possesses both specific capabilities tailored to bearing fault diagnosis and strong generalization abilities. Building on this, the model can be extended to address the PHM needs of a diverse range of multi-level objects. This includes devices of similar complexity like bearings, gears, actuators, and inertial groups, and even more complex systems like airplanes, satellites, ships, or even networks of satellites and systems networks.

To realize this transformation, challenges must be addressed at each step of constructing the Large Model. Its generalization capability mainly hinges on the pre-training and fine-tuning phases. Taking the bearing as an example, during the pre-training phase, the Large Model, by processing vast amounts of PHM data, can learn the intrinsic logic behind the perception, isolation, and judgment of bearing faults. This understanding then generalizes to capabilities like bearing assessment and prediction, as well as PHM capabilities for many other objects. However, the model post-pre-training lacks inference capabilities. To make the AI more comprehensively understand the nuances and connections between bearings, other similar-level objects, and even cluster-level entities, and to more rationally extrapolate the diagnosis learning to holistic PHM capabilities, presents a technological challenge that needs overcoming.

## 6.5 Interpretability and Traceability of PHM-LM Outputs

The research community has yet to delve deeply into the critical factors underpinning the superior capabilities of Large Model, as noted in the technical report by OpenAI on GPT-4 in 2023. As a result, understanding the sources of capabilities in the PHM-LM post-construction remains elusive. Ethical concerns in AI are mature for exploration, but the efficacy of results generated through a black-box approach remains questionable when applied to critical systems. The current solution involves increased human oversight and expert assessment, yet the traceability of outputs from future fully automated PHM generation tasks is drawing attention.

On another note, the PHM-LM's output capabilities cannot guarantee alignment with human preferences and values. Even with outstanding capabilities, there's a risk it could produce harmful, fictitious, or negatively impactful content. For instance, the phenomenon of hallucination refers to generated information conflicting with existing sources (intrinsic hallucination) or unverifiable by existing sources (extrinsic hallucination). This leads to the Large Model generating outputs that may seem credible but are entirely inconsistent with system characteristics, which can be catastrophic.

In the realm of LLMs, InstructGPT has designed effective fine-tuning methods, which include reinforcement learning techniques based on human feedback. This approach enhances an LLM's generalization capabilities on unseen tasks by providing human directives. While reinforcement learning techniques can be applied to the PHM-LM, enhancing the interpretability of the AI outputs remains a long-standing challenge in the Large Model research community.

## 6.6 Synergies Between Generalized PHM Methods based on Large Model and Existing Proprietary Methods

The paradigms discussed in this article encompass functions such as text generation, solution generation, capability generation, knowledge management, algorithm recommendation, assisted development, and validation assessments. These can lead to a generalized PHM series of methods based on the Large Model, while current specialized methods predominantly rely on non-AI means like expert experience, industry standards, manual code development, and human-conducted tests. While the PHM-LM clearly has advantages in intelligence, generalization, and generative capabilities, research indicates that when a Large Model excels in certain areas, it may face challenges in others(Yao et al., 2023). When there's a conflict between old and new knowledge, the catastrophic forgetting issue in NLP becomes

evident(Shinn et al., 2023; G. Wang et al., 2023). The hallucination phenomenon is also a major challenge for models like GPT, posing risks in the PHM field of generating harmful conclusions with adverse consequences. In a specific system, operational conditions, and service requirements, the model's generative expertise can be constrained; for critical tasks, the output capabilities might be far from expert knowledge, endangering the remaining life of the system and the safety of its operators. Therefore, even if a generalized PHM methodology is introduced, the existing specialized methodology should be retained and the two should operate in tandem. On the training front, the Large Model can be continually fine-tuned using specialized knowledge, employing experts to guide the model in learning inference methods for specific objects and scenarios, thus integrating specialized methods into the general model training and steadily approaching the desired outcomes. On the application front, one should not solely rely on the output from the general PHM methods but should implement solutions after thorough validation and assessment.

## 6.7 Distributed Privacy Training for PHM-LM Balancing Data Privacy and Sharing

Previously mentioned, considering the characteristics of PHM data such as privacy and dispersion, it's necessary to integrate vast resources and data within the PHM field. This process involves distributed training and subsequently data privacy issues. When training PHM on a large dataset within a single machine, constraints arise concerning data privacy security(S. Kim et al., 2023). Federated Learning (FL) is a machine learning framework that refers to the distribution of dispersed training data among various clients. These clients collaboratively train a single model under the coordination of a central server, embodying principles of centralized collection and data minimization, mitigating systematic privacy risks and costs(Q. Yang et al., 2019). By employing an FL approach, the PHM-LM can achieve data sharing and joint modeling on the premise of ensuring data legality and compliance, which also guarantees the effective implementation once the Large Model is deployed. However, using FL to train the PHM-LM faces several technical challenges in practical applications(Q. Wu et al., 2020), such as heterogeneity in storage, computation, and communication capabilities across clients, non-independent and identically distributed local data among clients, model heterogeneity issues based on varying application scenarios of different clients, potential hacks leading to privacy breaches, and so forth. Since FL technology is still in its nascent stages, further exploration is needed to assess its feasibility as a training framework for the PHM-LM and to find solutions for the associated challenges.

## 6.8 Summary

The PHM-LM, serving as a fundamental technological transformation and a novel means of achieving intelligence in the PHM industry, will confront numerous challenges across stages such as dataset construction, information representation architecture, structural design, training methods, deployment approaches, and operational workflows. Concurrently, there’s significant room for improvement in areas like generalizable performance, interpretability and traceability of output results, and synergy between general and specialized methods. Considering the multi-modality, multi-capability characteristics of the PHM field, alongside the requirements for generality and interpretability, this section proposes a fusion strategy integrating PHM with Large Model, edge-cloud collaboration, federated learning, and distributed training. This offers insights into addressing a broader range of technical challenges associated with the PHM-LM.

## 7. Conclusion

To address the challenges faced by the existing PHM and cater to the demands of future models, this paper, after analyzing the bottlenecks in the development of PHM technology and researching the evolution and advantages of Large Model technology, introduces the concept of a generative multi-modal PHM-LM. By combining the strengths of Large Model technology, the systems engineering process throughout the entire life cycle of PHM systems, and the problems faced by PHM, we present 3 prototypical innovative paradigms for advancing the research and application of the PHM-LM. This work delves into the technical process of transitioning from traditional LLMs to PHM-LM, proposing a specialized field application design for the PHM-LM. This disrupts conventional PHM design patterns, research and development methods, validation modes, and application methods, constructing a new technical system framework and ecosystem blueprint for the PHM-LM. This serves as a crucial reference and guide for realizing the generational leap and fundamental transformation in PHM technology – moving from traditional customization to generalization, from discriminative to generative, and from idealized conditions to real-world applications.

## Acknowledgement

This study was supported by the National Key R&D Program of China under Grant STI 2030—Major Projects (Grant No. 2021ZD0201300), the National Natural Science Foundation of China (Grant Nos. 61973011, 62103030, and 61903015), the Fundamental Research Funds for the Central Universities (Grant No. YWF-23-L-711), as well as the Capital Science & Technology Leading Talent Program (Grant No. Z191100006119029).

## Author Contributions

Conceptualization, L.T.; Data curation, L.M. and Z.Z.; Formal analysis; Funding acquisition, C.L. and C.W.; Methodology, L.T., S.L. and L.M.; Project administration, L.T., S.L. and L.M.; Writing-original draft, S.L., H.F.L., Q.H., G.N., Y.C., Y.W., B.L., W.W.Z., W.C.Z, and W.C.; Writing-review & editing, L.T., C.W., H.M.L., J.M., M.S., Y.C., Y.D. and D.S.; Validation, S.L.; Visualization, H.F.L. and Z.Z.; Investigation, X.S.; Supervision, C.L. All authors have read and agreed to the published version of the manuscript.